%% file: main_arxiv.tex
\documentclass{article}

\usepackage[nonatbib, final]{nips_2017}

\usepackage{nips_2017}

\usepackage[utf8]{inputenc} 
\usepackage[T1]{fontenc}    
\usepackage{hyperref}       
\usepackage{url}            
\usepackage{booktabs}       
\usepackage{amsfonts}       
\usepackage{nicefrac}       
\usepackage{microtype}      

\usepackage{graphicx}
\usepackage{color}
\usepackage{amsmath}
\usepackage{subfig}
\usepackage{multirow}
\usepackage{overpic}
\usepackage{wrapfig}

\title{On Nearest Neighbors in Non Local Means Denoising}

\author{
  I. Frosio \\
  NVIDIA \\
  USA \\
  \texttt{ifrosio@nvidia.com} \\
  \And
  J. Kautz \\
  NVIDIA \\
  USA \\
  \texttt{jkautz@nvidia.com}
}

\begin{document}

\maketitle

\begin{abstract}
To denoise a reference patch, the Non-Local-Means denoising filter processes a set of neighbor patches.
Few Nearest Neighbors (NN) are used to limit the computational burden of the algorithm. Here here we show analytically that the NN approach introduces a bias in the denoised patch, and we propose a different neighbors' collection criterion, named Statistical NN (SNN), to alleviate this issue.
Our approach outperforms the traditional one in case of both white and colored noise: fewer SNNs generate images of higher quality, at a lower computational cost. 
\end{abstract}

\input{Introduction}
\input{RelatedWork}

\input{Method}

\input{Results}
\input{Discussion}

\bibliography{references}
\bibliographystyle{acm}

\pagebreak
\rule{\textwidth}{3pt}
\vspace{-1cm}
\section*{
\center{\LARGE{On Nearest Neighbors in Non Local Means Denoising: Additional Material}}}
\hrulefill

\setcounter{section}{0}

\section{Prediction error of NN, SNN}

In this section we resort to a toy problem to compute analytically the bias and variance of the estimate of a noise-free patch, in case neighbor patches are collected through the traditional NN approach or trough the proposed SNN method. We also demonstrate numerically the advantage of SNN over NN for the toy problem in hand.

\subsection{Prediction error of NN}
\label{sec:prediction_error_of_NN}

\begin{figure*}
\begin{center}
\subfloat[]{\includegraphics[clip=true, trim = 2.5cm 0.5cm 2.5cm 0cm, width=0.48\textwidth]{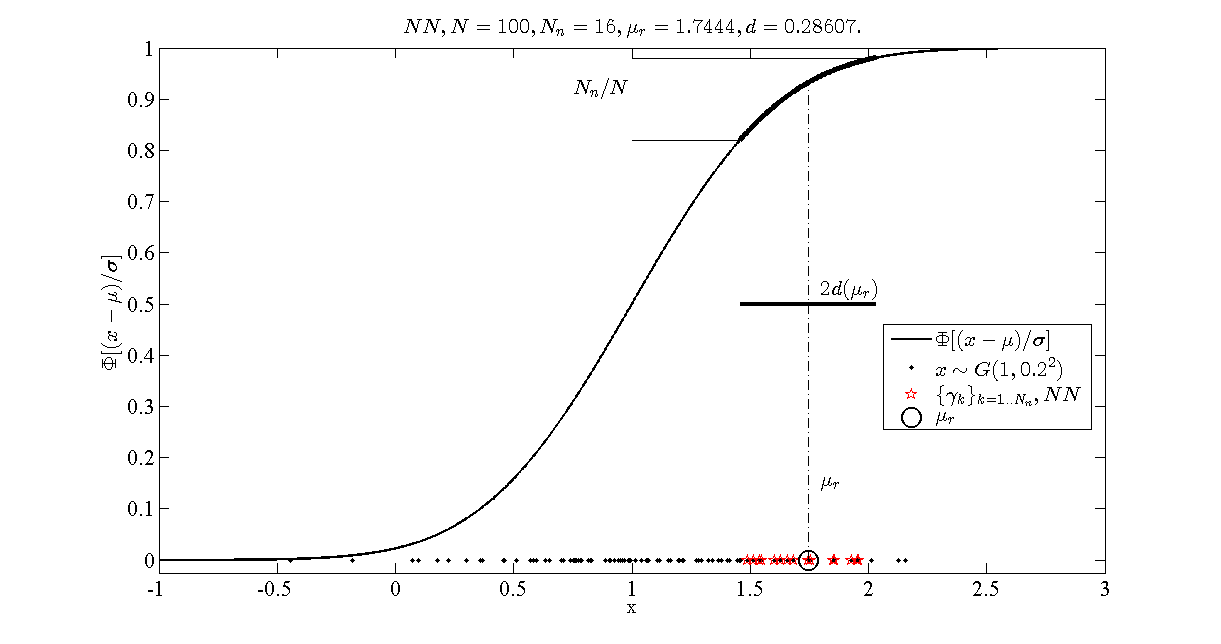}\label{fig:searchStrategies_a}}
\subfloat[]{\includegraphics[clip=true, trim = 2.5cm 0.5cm 2.5cm 0cm, width=0.48\textwidth]{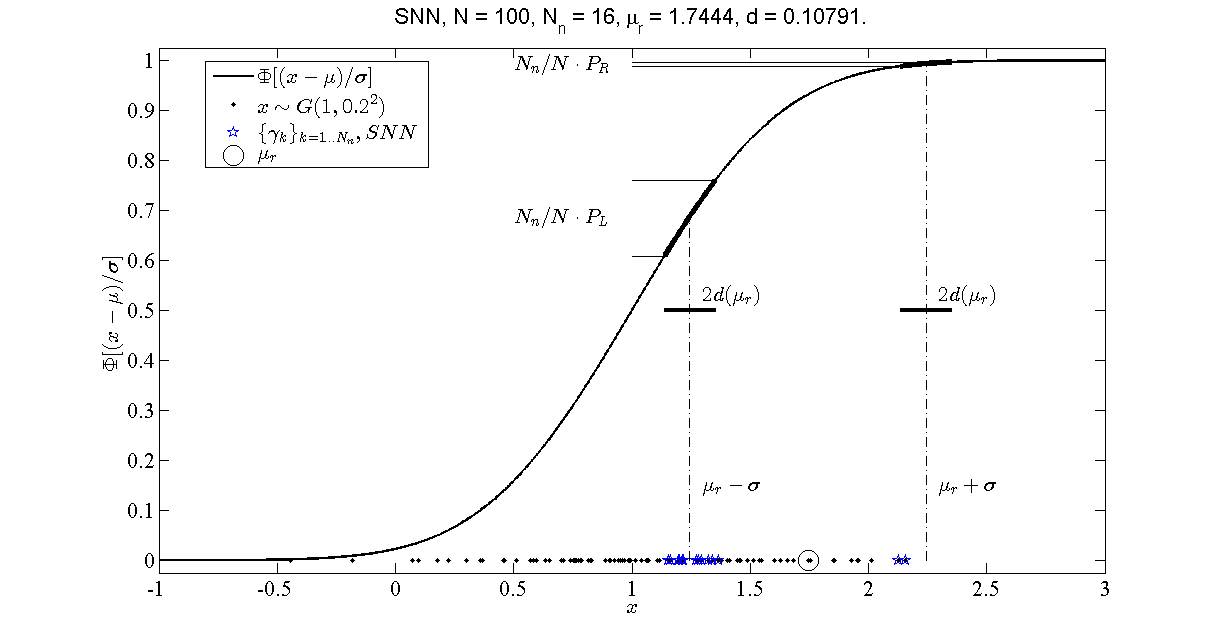}\label{fig:searchStrategies_b}}\\
\subfloat[]{\includegraphics[clip=true, trim = 2.5cm 0.5cm 2.5cm 0cm, width=0.48\textwidth]{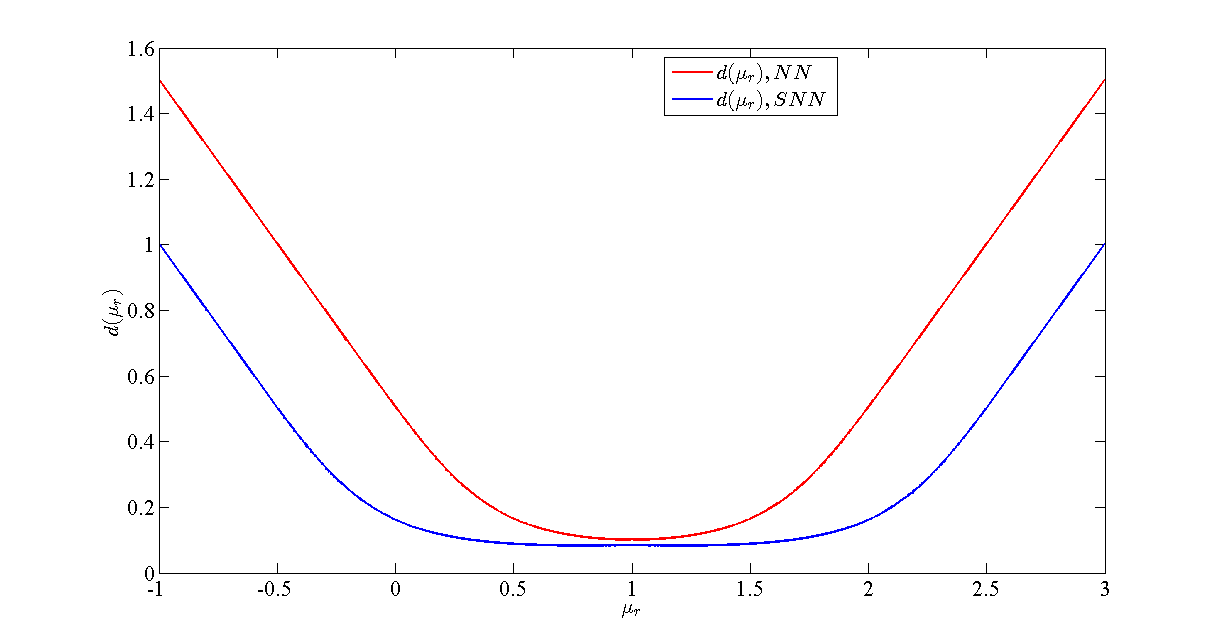}\label{fig:searchStrategies_c}}
\subfloat[]{\includegraphics[clip=true, trim = 2.5cm 0.5cm 2.5cm 0cm, width=0.48\textwidth]{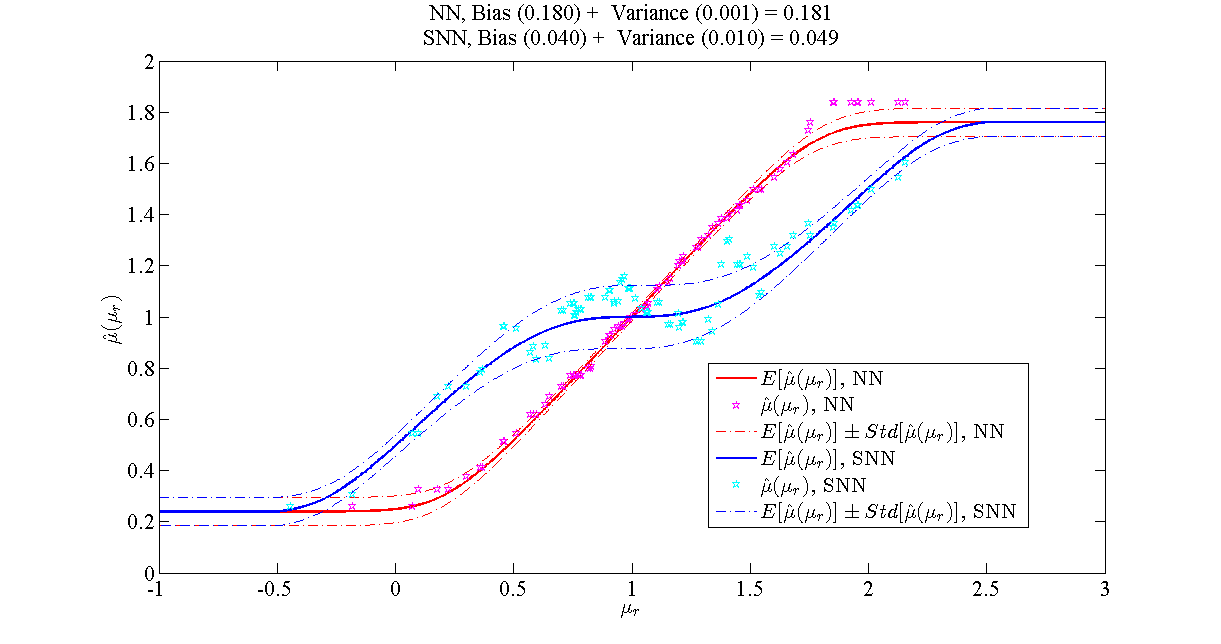}\label{fig:searchStrategies_d}}
\caption{Panels (a) and (b) show respectively the NN and SNN strategy to collect $N_n = 16$ neighbors $\{\gamma_k\}_{k=1..N_n}$ of $\mu_r$, for a $1 \times 1$ patch with noise-free value $\mu = 1$, corrupted by zero-mean Gaussian noise, $\sigma = 0.2$, from a set of $N = 100$ samples. In panels (a) and (b), $\mu_r = 1.7444$, far off $\mu=1$. Its 16 NNs are in its immediate vicinity (red, in a), ultimately leading to a bad estimate. The 16 SNNs (blue, in b) are on average closer to the actual $\mu$ leading to a better estimate. Panel (c) shows the neighbor search intervals; $d(\mu_r)$ is generally smaller for SNN, since the search for neighbors occurs in this case in two ranges, on the left and right of $\mu_r$. Panel (d) shows the expected value of the estimate, $\text{E}[\hat{\mu}(\mu_r)]$, and its  standard deviation, $\text{Std}[\hat{\mu}(\mu_r)]$, as a function of $\mu_r$; $\hat{\mu}(\mu_r)$ are specific estimates of $\mu$ obtained from the black samples $x$ in panels (a) and (b). SNN yields values closer to the actual $\mu=1$, even when the noisy patch $\mu_r$ is far off the actual noise-free value.}
\label{fig:searchStrategies}
\end{center}
\vspace{-1.5em}
\end{figure*}

NLM denoising is a three step procedure: i) for each patch, the neighbors are identified; ii) neighbors are averaged trough a weighted average procedure); iii) denoised patches are aggregated (since patches are partially overlapping, multiple estimates are averaged for each pixel).
Collecting NN neighbors introduces a prediction error in step ii).
To demonstrate this, we consider a $1 \times 1$ patch with noise-free value $\mu$, and assume Gaussian noise\footnote{Our reasoning  for a single-pixel patch easily generalizes to a larger patch, where each pixel is corrupted by zero-mean Gaussian noise.} with variance $\sigma^2$.
We assume that $N$ noisy replicas of the patch are available (black samples in Fig.~\ref{fig:searchStrategies_a}).
For each reference patch, $\mu_r$, we collect $N_n$ neighbors, $\{\gamma_k\}_{k=1.. N_n}$; $\gamma_k$ and $\mu_r$ are Gaussian random variables with average $\mu$ and variance $\sigma^2$, indicated by $G(\mu, \sigma^2)$, with pdf $1 / \sigma \cdot \phi \left[ (x-\mu) / \sigma \right]$, where $\phi(x) = 1 / \sqrt{2 \pi} \cdot e^{-0.5 \cdot x^2}$.
We define $d = d( \mu_r)$ the range in which we are likely to find the $N_n$ nearest neighbors of $\mu_r$ (red stars in Fig.~\ref{fig:searchStrategies_a}). More specifically, $d(\mu_r)$ satisfies:
%
\begin{eqnarray}
\frac{1}{\sigma} \int_{\mu_r - d(\mu_r)}^{\mu_r + d(\mu_r)}{\phi \left(\frac{x-\mu}{\sigma} \right) dx} \!\!&=&\!\! \nonumber \\
\Phi[\frac{\mu_r + d(\mu_r) -\mu}{\sigma}] - \Phi[\frac{\mu_r - d(\mu_r) -\mu}{\sigma}] \!\!&=&\!\! \frac{N_n}{N}, 
\label{eq:d}
\end{eqnarray}
%
where $\Phi(x) = 1 / 2 \cdot [1 + \text{erf}(x/\sqrt{2})]$, and $\Phi[(x-\mu) / \sigma]$ is the cdf of a Gaussian random variable with mean $\mu$ and variance $\sigma^2$. A closed form solution for $d(\mu_r)$ in Eq.~(\ref{eq:d}) does not exist, but $d(\mu_r)$ can be computed numerically, for instance, using a bracketing technique (see Fig.~\ref{fig:searchStrategies_c}). From $d(\mu_r)$, we compute the expected value of the average of the $N_n$ nearest neighbors around $\mu_r$, which is $E[\hat{\mu}(\mu_r)]$. This is equivalent to step ii) of NLM, neglecting the weights in the weighted average. Since $\hat{\mu}(\mu_r)$ is a truncated Gaussian variable, bounded by $\mu_r \pm d(\mu_r)$, its expected value and variance are:
\begin{eqnarray}
\alpha &=& \frac{[\mu_r - d(\mu_r)]}{\sigma}, \ \beta = \frac{[\mu_r + d(\mu_r)]}{\sigma} \nonumber
\label{eq:alphaBetaNN} \\
\text{E}[\hat{\mu}(\mu_r)] &=& \mu - \sigma \cdot \frac{\phi(\beta) - \phi(\alpha)}{\Phi(\beta) - \Phi(\alpha)}
\label{eq:ENN} \\
\text{Var}[\hat{\mu}(\mu_r)] &=& \sigma^2 \cdot \{1 - \frac{\beta \cdot \phi(\beta) - \alpha \cdot \phi(\alpha)}{\Phi(\beta) - \Phi(\alpha)} + \nonumber \\ & & - [\frac{\phi(\beta) - \phi(\alpha)}{\Phi(\beta) - \Phi(\alpha)}]^2\} / N_n.
\label{eq:VarNN}
\end{eqnarray}
Fig.~\ref{fig:searchStrategies_d} shows $\text{E}[\hat{\mu}(\mu_r)]$ and $\text{Std}[\hat{\mu}(\mu_r)]$ for the case $\mu = 1$, $\sigma = 0.2$, when $N_n = 16$ neighbors are to be collected from a total of $N = 100$ samples. Since the NN neighbors $\{\gamma_k\}_{k=1..N_n}$ lie close to the noisy reference patch, $\mu_r$, each estimate $\hat{\mu}(\mu_r)$ is biased towards $\mu_r$. The bias grows almost linearly with $\mu-\mu_r$ and it saturates at approximately $5 \sigma$ of distance from $\mu$, as the set of neighbors becomes stable (since $N$ is finite, the same 16 samples are found in the tail of the distribution). We compute the expected prediction error for estimating $\mu$ through $\hat{\mu}(\mu_r)$ by integrating the bias and variance terms as:
\begin{eqnarray}
MSE &=& \int_{-\infty}^{+\infty} \!\!\!\!\!\!\! \{[\text{E}[\hat{\mu}(\mu_r)] - \mu]^2 + \text{Var}[\hat{\mu}(\mu_r)]\} \cdot \nonumber \\ & & \,\,\,\,\,\, \frac{1}{\sigma} \cdot \phi[\frac{(\mu_r-\mu)}{\sigma}] d\mu_r, 
\label{eq:MSE}
\end{eqnarray}
and it is equal to 0.180 (bias) + 0.001 (variance) = 0.181 for the case in Fig. \ref{fig:searchStrategies_a}. In practice, when NN neighbors are collected, samples with correlated noise are likely to be chosen. We call this the \emph{noise-to-noise} matching problem: averaging NNs patches may amplify small correlations between the noise, without canceling it. In the context of image denoising, the \emph{noise-to-noise} matching problems shows up as residual, colored noise in the filtered image (see the images filtered with NLM$^{16}_{0.0}$ in the main paper).


\subsection{Statistics of patch distance}
\label{sec:statistics_of_patch_distance}
Before illustrating the SNN method to collect the neighbors of a reference patch, we analyze the statistical distribution of the distance between multi-dimension patches (similarly to~\cite{Wu13}).
This shows that, in presence of noise, the expected distance between the reference patch and its neighbors is not zero, which is the main observation behind the proposed SNN approach.

The search for similar patches is performed by computing the distance between $\boldsymbol{\mu_r}$ and patches of the same size, but in different positions in the image. If, apart from noise, the  reference patch $\boldsymbol{\mu_r}$ and its neighbor $\boldsymbol{\gamma_k}$ are two replica of the same patch, we get:
\begin{equation}
\delta^2 \left(\boldsymbol{\mu_r}, \boldsymbol{\gamma_k} \right)
=
(2 \sigma^2 / P) \cdot \sum_{i = 0}^{P-1}{G(0, 1)^2}.
\label{eq:distance2}
\end{equation}

Since the sum of $P$ squared normal variables has a $\chi_P^2$ distribution with $P$ degrees of freedom, we have $\delta^2~\left(\boldsymbol{\mu_r},~\boldsymbol{\gamma_k}~\right)~\sim~(2~\sigma^2~/~P)~\cdot~\chi_P^2$, and therefore:
\begin{equation}
E[\delta^2\left(\boldsymbol{\mu_r}, \boldsymbol{\gamma_k} \right)] = 2 \sigma^2.
\label{eq:expectedSquaredDistance}
\end{equation}

Thus, for two noisy replicas of the same patch, the expected squared distance is not zero. This  has already been noticed and effectively employed since the original NLM paper~\cite{Bua05} to compute the weights of the patches in the weighted average (see Eq. (\ref{eq:weight}) in the main paper), giving less importance to patches at a squared distance larger than $2 \sigma^2$, or to build a better weighting scheme, as in ~\cite{Wu13}.
Nonetheless, to the best of our knowledge, it has never been employed as a driver for the selection of the neighbor patches, as we do here.

We can also establish a connection between the toy problem at hand and matching patches in general. In fact, for our toy problem, we can apply Fischer's approximation ($\sqrt{ 2 \chi_P^2} \cong G(\sqrt{2P-1}, 1)$) to a single-pixel patch ($P = 1$), and get:
\begin{equation}
\delta \left(\boldsymbol{\mu_r}, \boldsymbol{\gamma} \right) \cong \sigma \cdot \sqrt{2P-1} + G(0, \sigma^2) = \sigma + G(0, \sigma^2),
\label{eq:expectedDistance}
\end{equation}
showing that the expected distance between two $1 \times 1$ noisy patches is approximately $\sigma$.

\subsection{SNN}
\label{sec:SNN}
As an alternative to NN, and mostly inspired by Eq. (\ref{eq:expectedSquaredDistance}), we propose collecting neighbors whose squared distance from the reference patch is instead close to its expectation. Thus SNNs are the patches $\{\boldsymbol{\gamma}_k\}_{k=1.. N_n}$ that minimize:
\begin{equation}
|\delta^2 \left(\boldsymbol{\mu_r}, \boldsymbol{\gamma}_k \right) - o \cdot 2\sigma^2|,
\label{eq:SNNs}
\end{equation}
where we have introduced an additional offset parameter $o$, that allows to continuously move from the traditional NN approach ($o=0$) to the SNN approach ($o=1$). We assume $o = 1$ for now.

%

\subsection {Prediction error of SNN}
\label{sec:prediction_error_of_SNN}

To compare the prediction error of NN and SNN in our toy problem, we compute $\text{E}[\hat{\mu}(\mu_r)]$ and $\text{Var}[\hat{\mu}(\mu_r)]$, when SNN neighbors of $\mu_r$ are collected, \emph{i.e.}, when samples are close to $\mu_r \pm o \cdot \sigma$\footnote{Notice that in our toy problem we collect the samples based on their expected distance from the noisy reference $\mu_r$, computed thorugh Eq. (\ref{eq:expectedDistance}), while in SNN we use the squared distance. This simplification makes the toy problem mathematically manageable, without changing the intuition behind SNN.}. Fig.~\ref{fig:searchStrategies_b} illustrates the sampling of SNN neighbors. Sampling potentially occurs on both sides of $\mu_r$, with a different chance of collecting neighbors on each side. The interval $d(\mu_r)$ where we expect to find the neighbors, has now to satisfy:
\begin{eqnarray}
\begin{array}{r}
\Phi\{[\mu_r - o \cdot \sigma + d(\mu_r) -\mu] / \sigma\} + \\ - \Phi\{[\mu_r - o \cdot \sigma - d(\mu_r) -\mu] / \sigma\} + \\
+ \Phi\{[\mu_r + o \cdot \sigma + d(\mu_r) -\mu] / \sigma\} + \\ - \Phi\{[\mu_r + o \cdot \sigma - d(\mu_r) -\mu] / \sigma\} = N_n / N,
\end{array}
\label{eq:dsnn}
\end{eqnarray}
where the first two rows represent the probability to sample one neighbor in the left interval ($P_L$ in Fig.~\ref{fig:searchStrategies_b}), whereas  the term in the third and fourth row is for the right interval ($P_R$ in Fig.~\ref{fig:searchStrategies_b}). Since $P_L > P_R$ here, the estimate $\hat{\mu}(\mu_r)$ will be likely moved towards the left, decreasing the prediction error. As for the NN case, $d(\mu_r)$ can be estimated through Eq.~(\ref{eq:dsnn}) using a bracketing technique. Fig.~\ref{fig:searchStrategies_c} shows the search intervals $d(\mu_r)$ for a Gaussian random variable, $\mu = 1$, $\sigma = 0.2$, when $N_n = 16$ and $N = 100$. Compared to the NN search range, the SNN one is smaller as it generally includes two search intervals. These two search ranges may collapse into one when $\mu_r$ belongs to the tales of the distribution, providing the same result as NN in this (rare) case.

The expected value of the average of the SNNs is the expected value of the mix of two truncated Gaussian random variables, on the left and right side of~$\mu_r$. After a few simplifications and using Eqs.~(\ref{eq:ENN}~-~\ref{eq:VarNN}) to compute the expected values ($\text{E}_L$,  $\text{E}_R$) and variances ($\text{Var}_L$, $\text{Var}_R$) on the left and right side of $\mu_r$, we have:
\begin{eqnarray}
\alpha_L &=& [\mu_r -o \cdot \sigma - d(\mu_r) - \mu] / \sigma, \nonumber \\
\beta_L &=&
[\mu_r -o \cdot \sigma + d(\mu_r) - \mu] / \sigma \nonumber \\ 
\alpha_R &=& [\mu_r +o \cdot \sigma - d(\mu_r) - \mu] / \sigma, \nonumber \\
\beta_R &=& [\mu_r +o \cdot \sigma + d(\mu_r) - \mu] / \sigma \nonumber \\ 
P_L &=& \Phi(\beta_L) - \Phi(\alpha_L), \ P_R = \Phi(\beta_R) - \Phi(\alpha_R) \nonumber \\
\text{E}[\hat{\mu}(\mu_r)] &=& (P_L E_L + P_R E_R) / (P_L + P_R)
\label{eq:ESNN}\\
\text{Var}[\hat{\mu}(\mu_r)] &=& \{\frac{P_L (E_L^2 + \text{Var}_L) + P_R(E_R^2 + \text{Var}_R)}{P_L + P_R}+\nonumber \\
& &- \text{E}[\hat{\mu}(\mu_r)]\} / N_n.
\label{eq:VarSNN}
\end{eqnarray}
%

Eqs. (\ref{eq:dsnn}), (\ref{eq:ESNN}), and (\ref{eq:VarSNN}) boil down to the simple NN case in the case of overlapping intervals.
Fig.~\ref{fig:searchStrategies_d} shows $\text{E}[\hat{\mu}(\mu_r)]$ and $\text{Std}[\hat{\mu}(\mu_r)]$ for the SNN case. The overall bias and variance for SNN are computed by integration as in Eq. (\ref{eq:MSE}), and they are respectively equal to 0.040 and 0.010, for a total MSE of 0.49. Compared to NN, SNN slightly increases the variance of $\hat{\mu}(\mu_r)$, but it drastically decreases the bias of the estimate, especially for those points close to $\mu$ (which are, by the way, the most frequent).
In practice, the SNN criterion looks for similar patches having orthogonal realizations of the noise.
This minimizes the \emph{noise-to-noise} matching issue and it leads to more effective noise cancellation, at the price of a slightly higher variance of $\hat{\mu}(\mu_r)$.

\section{Quantitative evaluation of the SNN approach for NLM denoising}
\label{sec:QuantiativeEvaluation}

We report here the extensive quantitative evaluation of the results achieved with the proposed SNN approach applied to NLM denoising, in the case of white Gaussian noise, colored noise, and in the case of a ``real'' image. Such evaluation has not been reported in the main paper for reasons of space.

\subsection{White Gaussian noise}

We test the effectiveness of the SNN schema on the Kodak image dataset~\cite{Pon13}.
We first consider denoising in case of white, zero-mean, Gaussian noise with standard deviation $\sigma = \{5, 10, 20, 30, 40\}$, and evaluate the effect of the offset parameter $o$ and number of neighbors $N_n$.
With NLM$^{N_n}_o$ we indicate NLM with $N_n$ neighbors and an offset $o$. 
Each image is processed with NLM$^{16}_o$ and for $o$ ranging from $ o = 0$ (\emph{i.e.}, the traditional NN strategy) to $o = 1$;
NLM$^{361}_{.0}$ and NLM$^{900}_{.0}$ indicate the original NLM, using all the patches in the search window, respectively for $\sigma < 30$ and $\sigma \geq 30$.
The patch size, the number of negihbors $N_n$, and the filtering parameter $h$ are the optimal ones in~\cite{Bua05}.

We resort to a large set of image quality metrics to evaluate the filtered images: PSNR, SSIM\cite{Wan04}, MSSSIM\cite{Wan03}, GMSD\cite{Xue14}, FSIM and FSIM$_C$\cite{Zha11b}. PSNR is proportional to the average squared error in the image; SSIM and MSSSIM take inspiration from the the human visual system, giving less importance to the residual noise close to edges. GMSD is a perceptually-inspired metric that compares the gradients of the filtered and the reference images. FSIM is the one that correlates better with the human judgement of the image quality, and takes into consideration the structural features of the images; FSIM$_C$ also takes into account the color component.

Table~\ref{table:NLM_sigmaall} shows the average image quality metrics measured on the Kodak dataset, whereas visual inspection of Fig.~\ref{fig:results_kodak} in the main paper reveals the correlation between these metrics and the artifacts introduced in the filtered images. The PSNR decreases when reducing the number of NN neighbors (see the first two rows for each noise level in Table~\ref{table:NLM_sigmaall}), mostly because of the residual, colored noise left by NLM$^{16}_{.0}$ in the flat areas (like the skin of the girl and the wall with the fresco in Fig.~\ref{fig:results_kodak} in the main paper); SSIM and MSSSIM show a similar trend, while GMSD, FSIM, and FSIM$_C$ improve when the number of NN neighbors is reduced to 16. The improvement of these three perceptually-based image quality metrics is associated with the reduction of the NLM bias in the case of a small set of neighbors, clearly explained in~\cite{Duv11}. At visual inspection, edges and structures are better preserved, (see the textile pattern and the sun rays in Fig.~\ref{fig:results_kodak} in the main paper). Table~\ref{table:NLM_sigmaall} shows the PSNR consistently increasing with the offset $o$, coherently with the decrease of the prediction error measured in our toy problem from $o = 0.0$ (NN) to $o = 1.0$ (SNN). Visual inspection (Fig.~\ref{fig:results_kodak} in the main paper) confirms that SNN removes more noise in the flat areas (\emph{e.g.}, the girl's skin, the wall surface) compared to NN. For $o = 1.0$ the PSNR approaches that achieved by traditional NLM, which requires more neighbor patches and therefore a higher computational time. Even more interestingly, SSIM and MSSSIM are maximized for $o = 0.9$, whereas the best GMSD, FSIM and FSIM$_C$ are obtained for $o = 0.65$ or $o = 0.8$. Even if none of these metrics is capable of perfectly measuring the quality perceived by a human observer, they consistently suggest that filtering with the SNN approach generates images of superior, perceptual quality compared to the traditional NLM filtering.
To summarize, sharp edges are reconstructed well by all the approaches. Low-contrast edges (\emph{e.g.}, the texture of the textile, the sun rays and the small details in the fresco), are well preserved by the NN approach, which achieves better GMSD, FSIM and FSIM$_C$ compared to traditional NLM, but NLM$^{16}_0$ also leaves higher, colored noise in the image (low PSNR, SSIM and MSSSIM). These same low-contrast edges are oversmoothed by NLM$^{361}_{0.0}$ and by SNN when $o = 1$ (NLM$^{361}_{1.0}$), whereas SNN with an offset $o = 0.8$ (NLM$^{361}_{0.8}$) achieves the best compromise between preserving low-contrasted details in the image and effectively smoothing the flat areas, coherently with the numerical evaluation in Table \ref{table:NLM_sigmaall}. The proposed SNN strategy always outperforms NN (when using the same number of patches) in the image quality metrics, with the exception of very low noise ($\sigma=5$), where GMSD, FSIM and FSIM$_C$ are constant for $.0 < o < 0.65$. Quite reasonably, the advantage of SNN over NN is larger for a large $\sigma$: notice in fact that, for $\sigma \rightarrow 0$, NN and SNN converge to the same algorithm.


\input{Tables/NLM_sigmaall}

\begin{figure*}
\centering
\begin{tabular}{c}
\hspace{-0.8cm}
\includegraphics[width=0.24\textwidth]{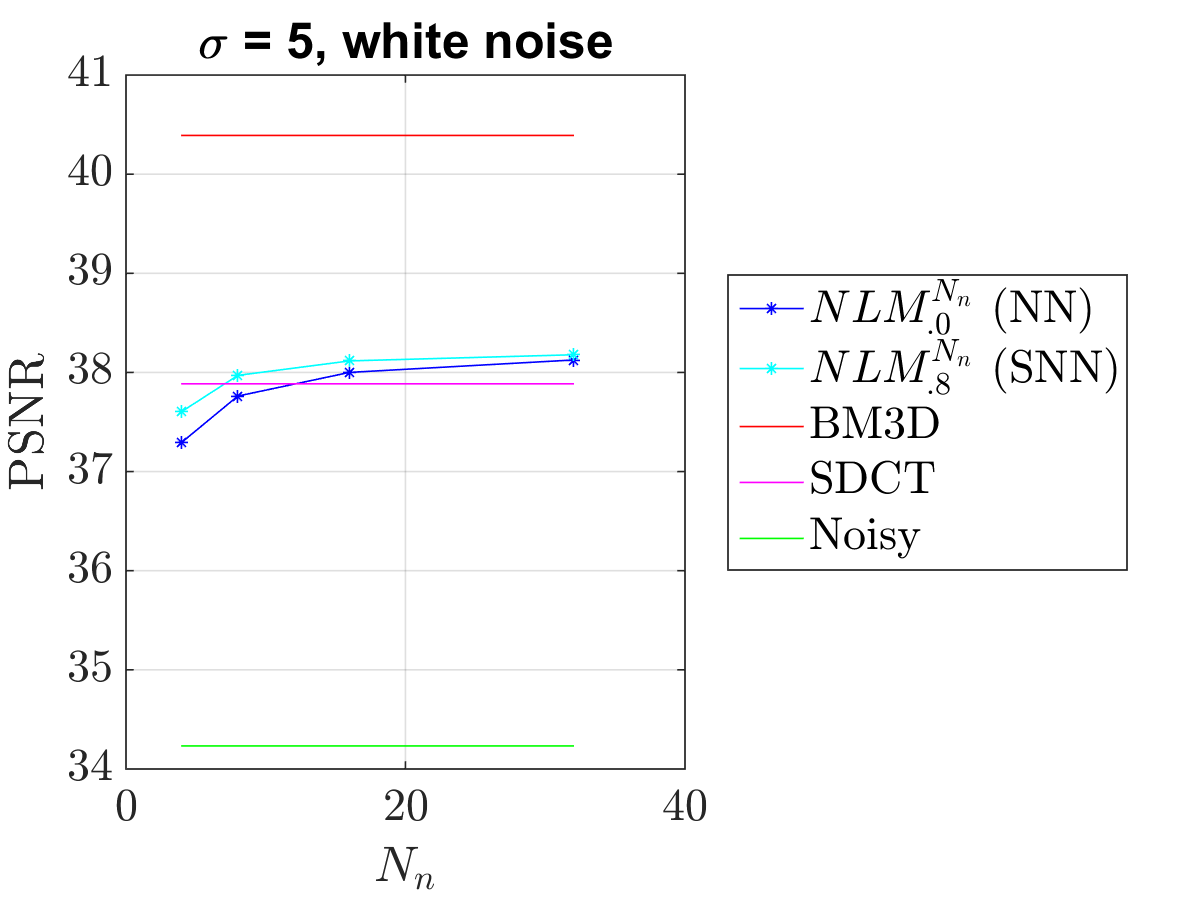}
\hspace{-2.0cm}
\includegraphics[width=0.24\textwidth]{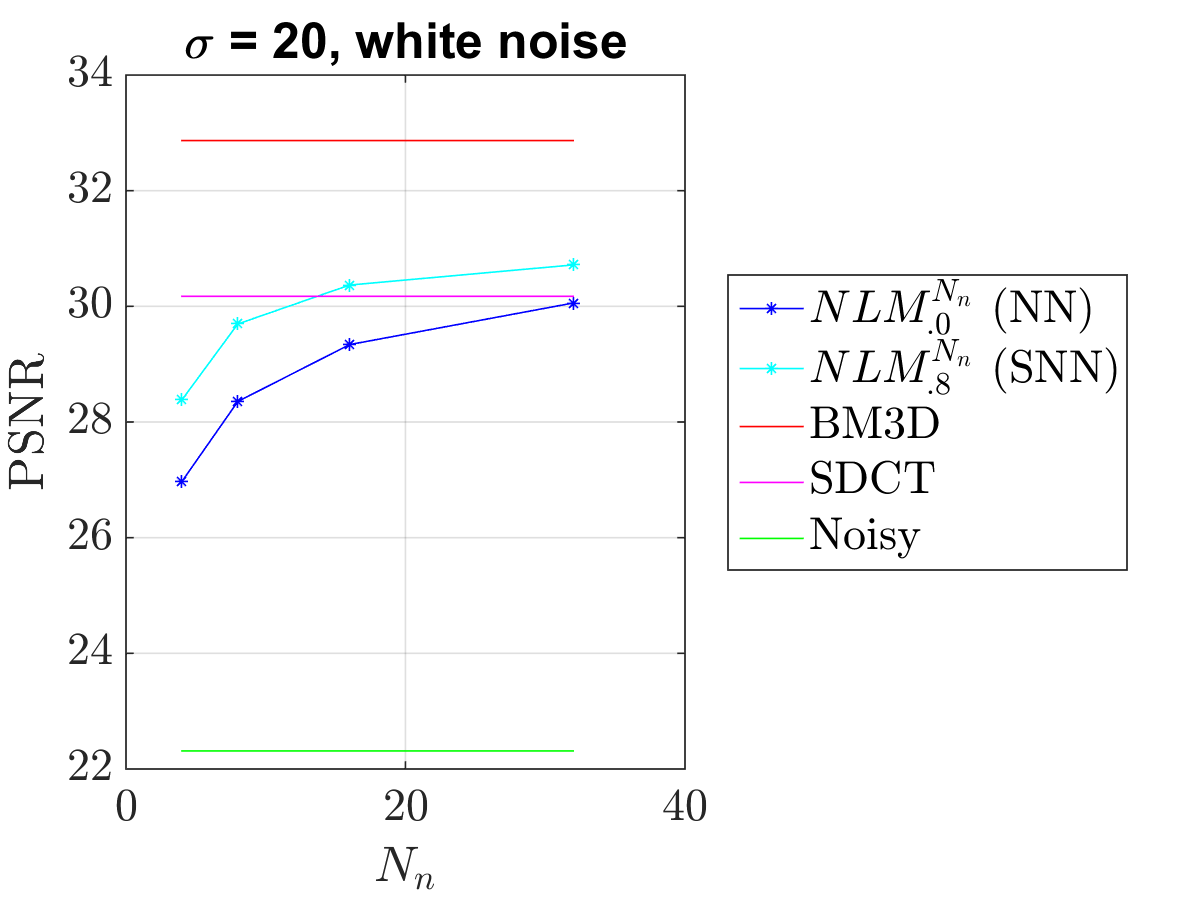}
\hspace{-2.0cm}
\includegraphics[width=0.24\textwidth]{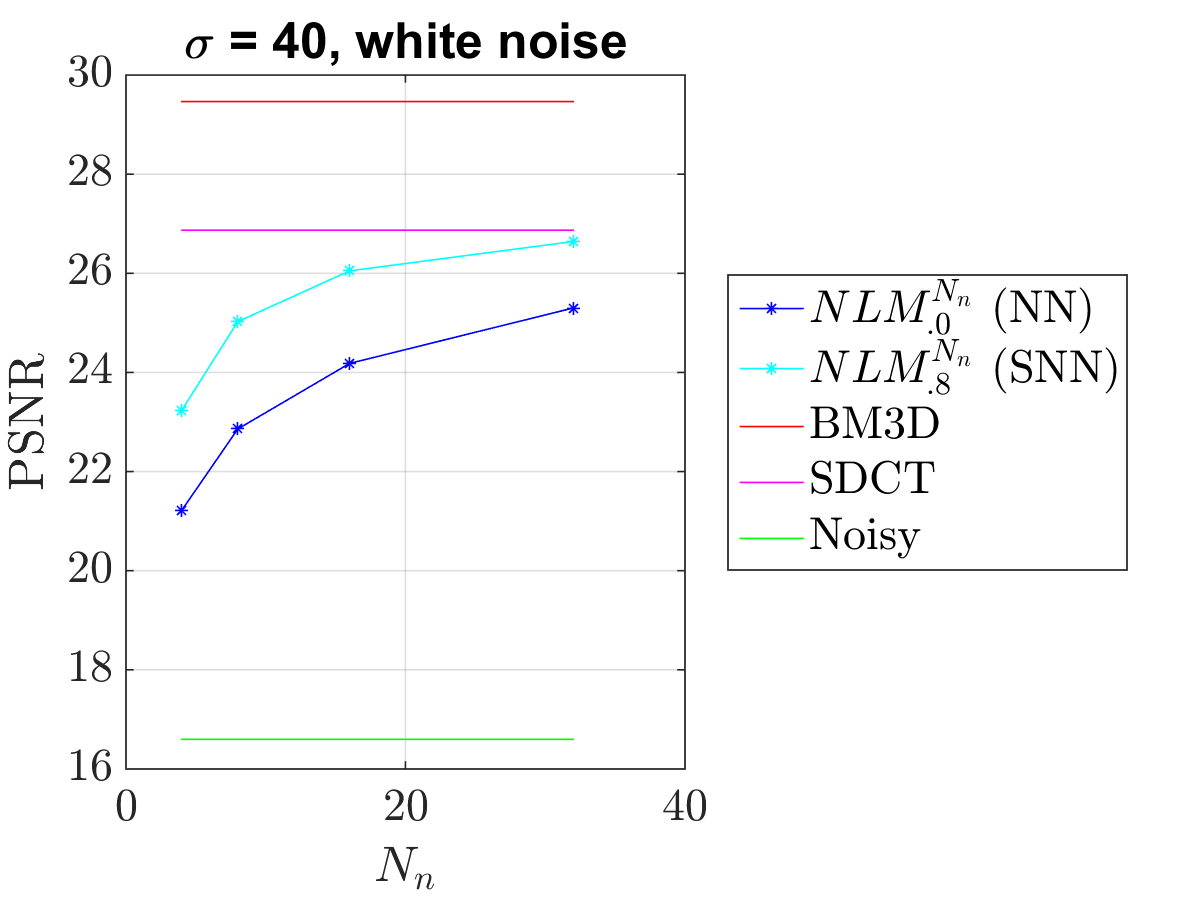}
\hspace{-2.0cm}
\includegraphics[width=0.24\textwidth]{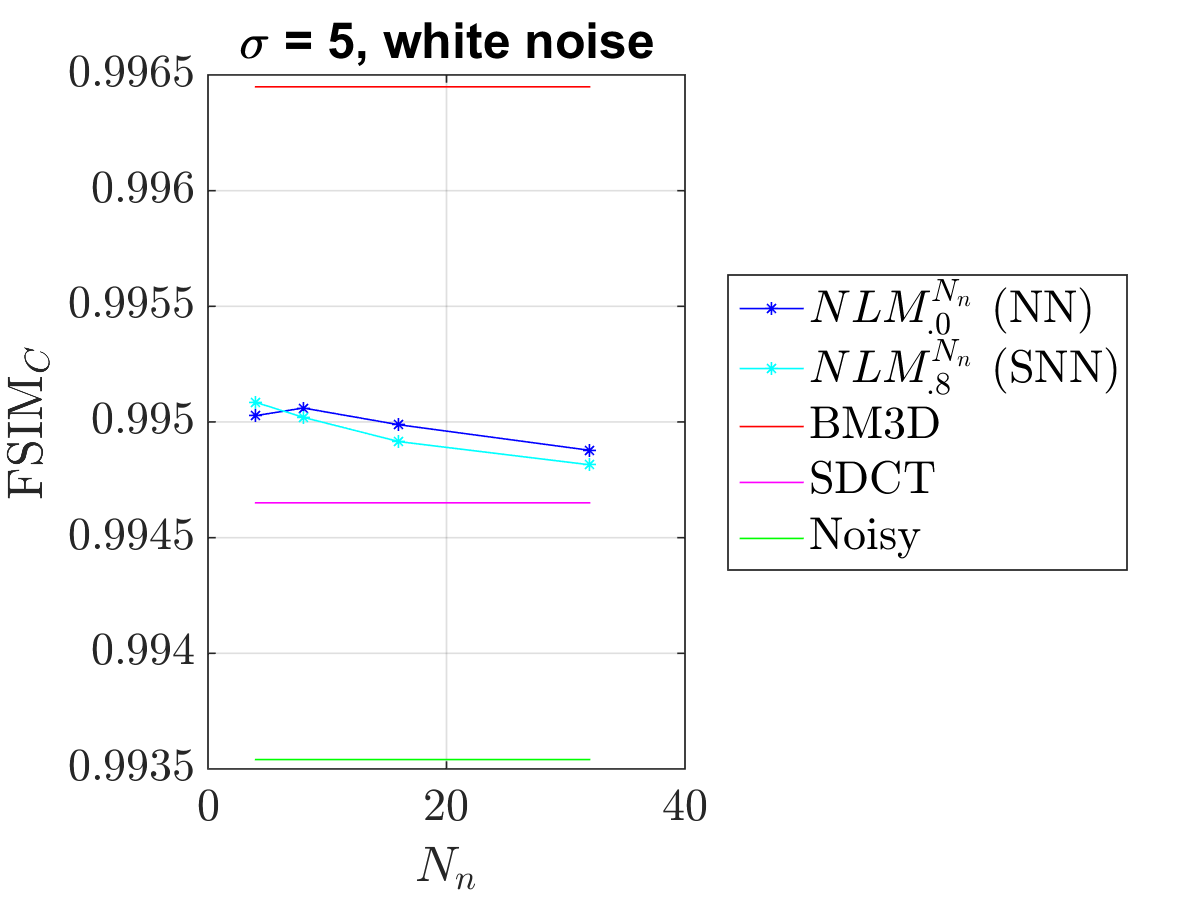}
\hspace{-2.0cm}
\includegraphics[width=0.24\textwidth]{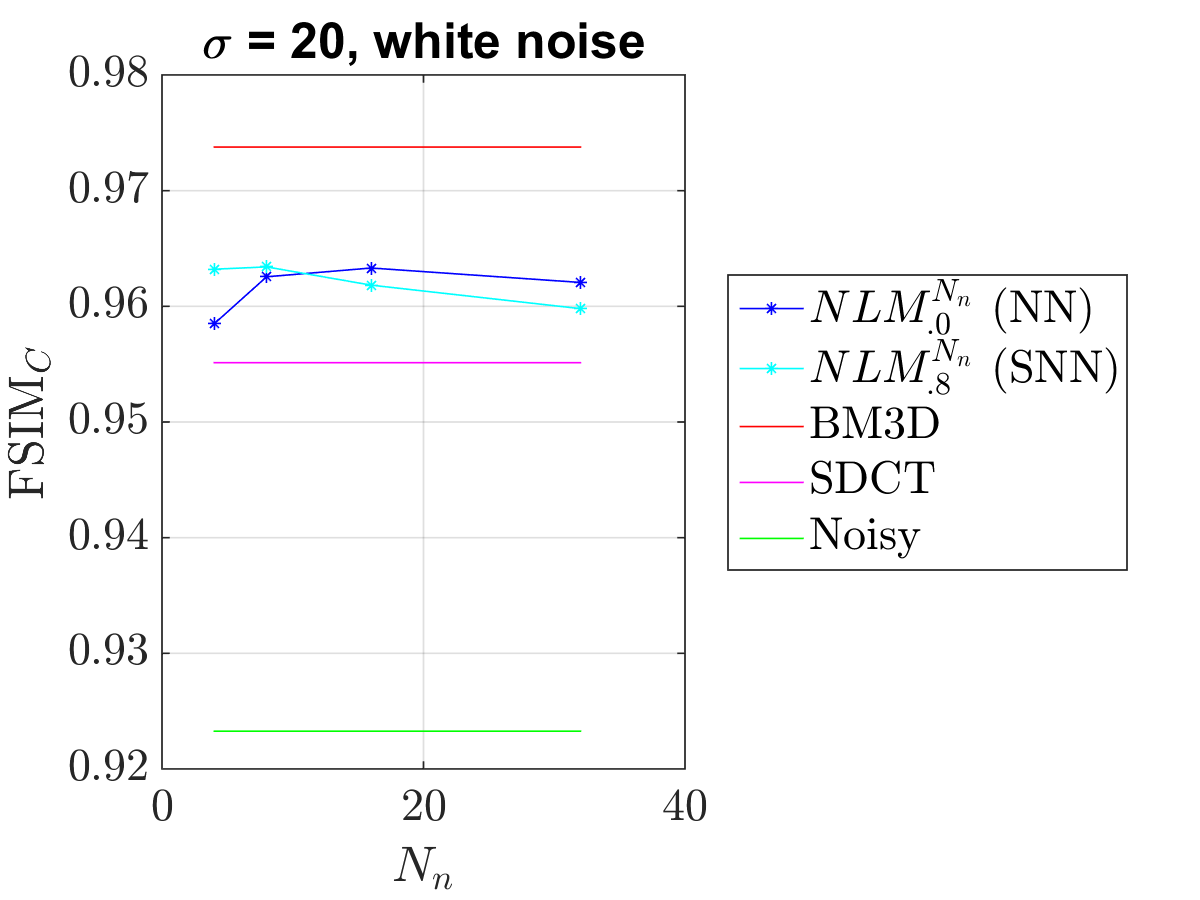}
\hspace{-2.0cm}
\includegraphics[width=0.24\textwidth]{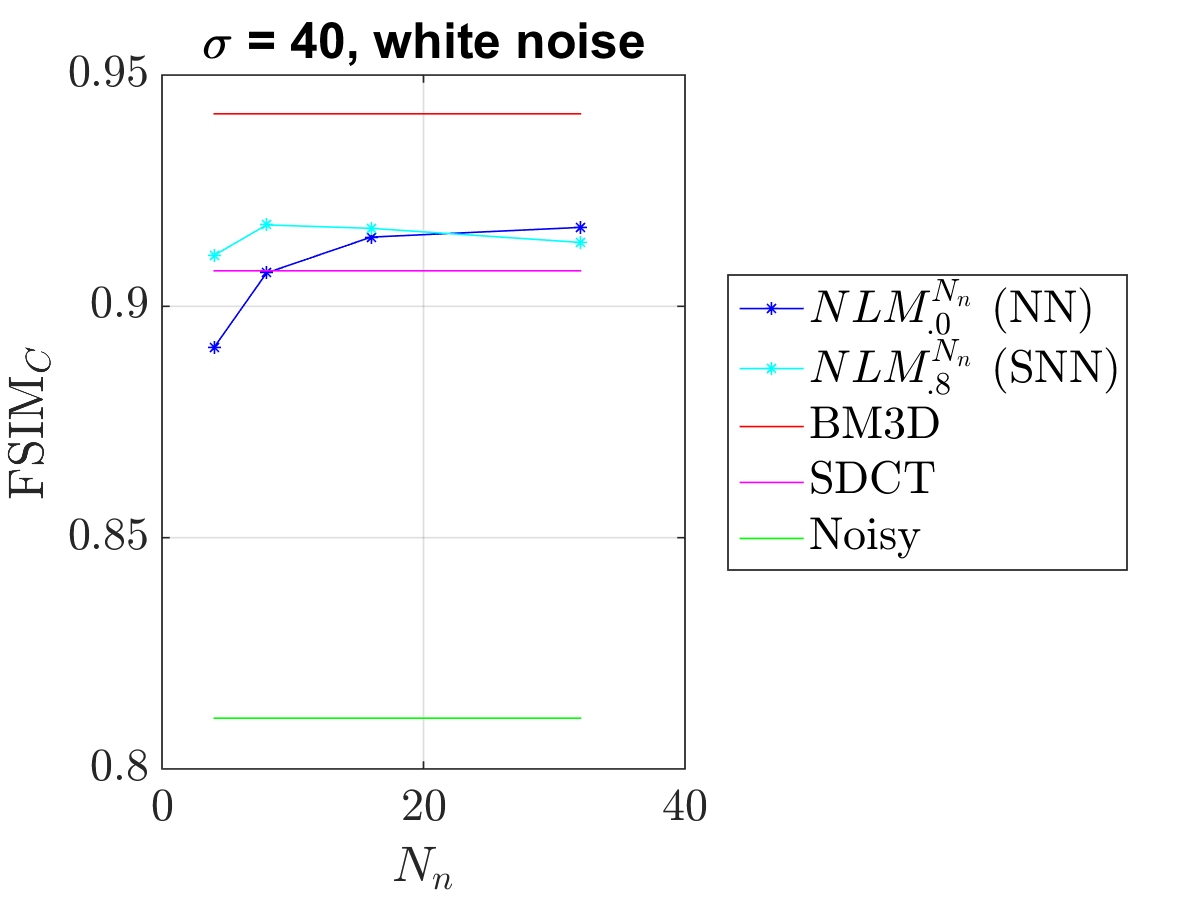}\\
\hspace{-0.8cm}
\includegraphics[width=0.24\textwidth]{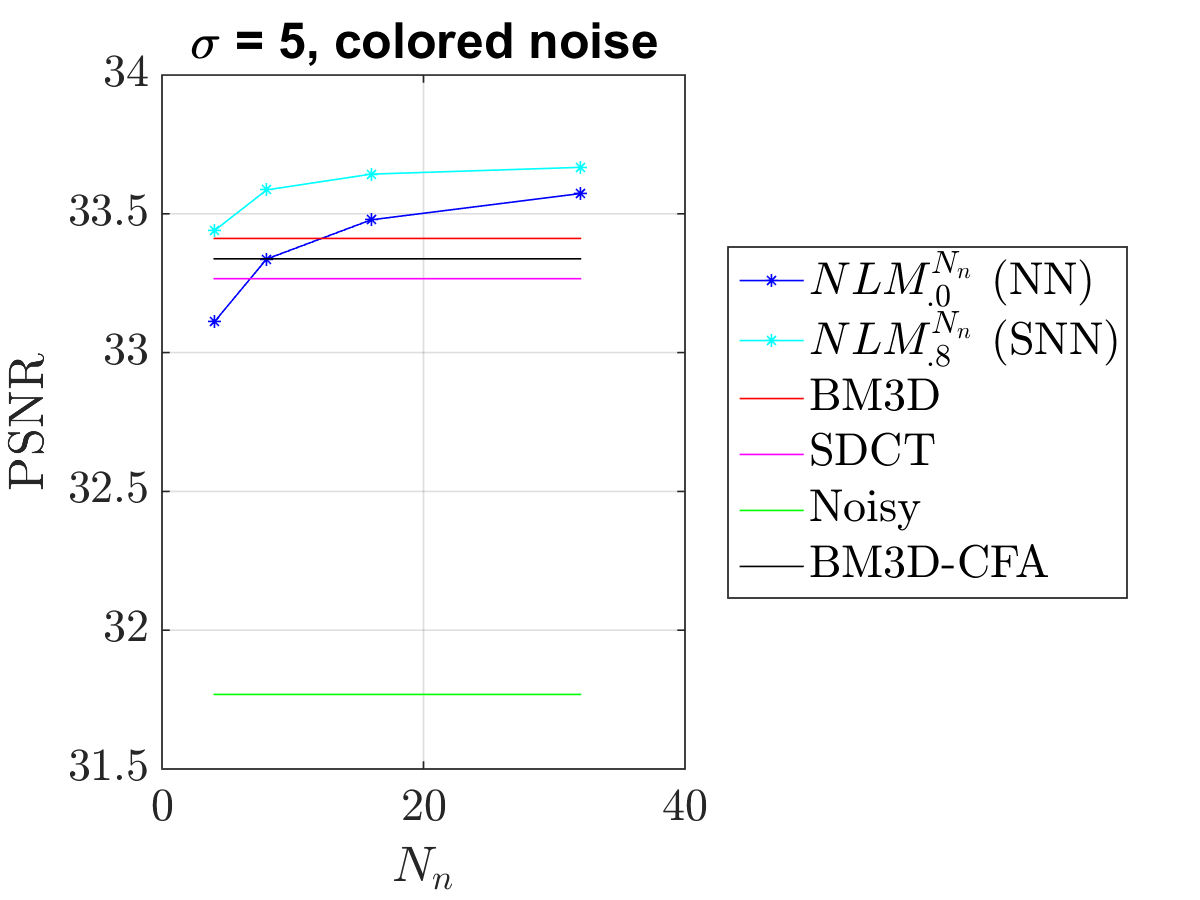}
\hspace{-2.0cm}
\includegraphics[width=0.24\textwidth]{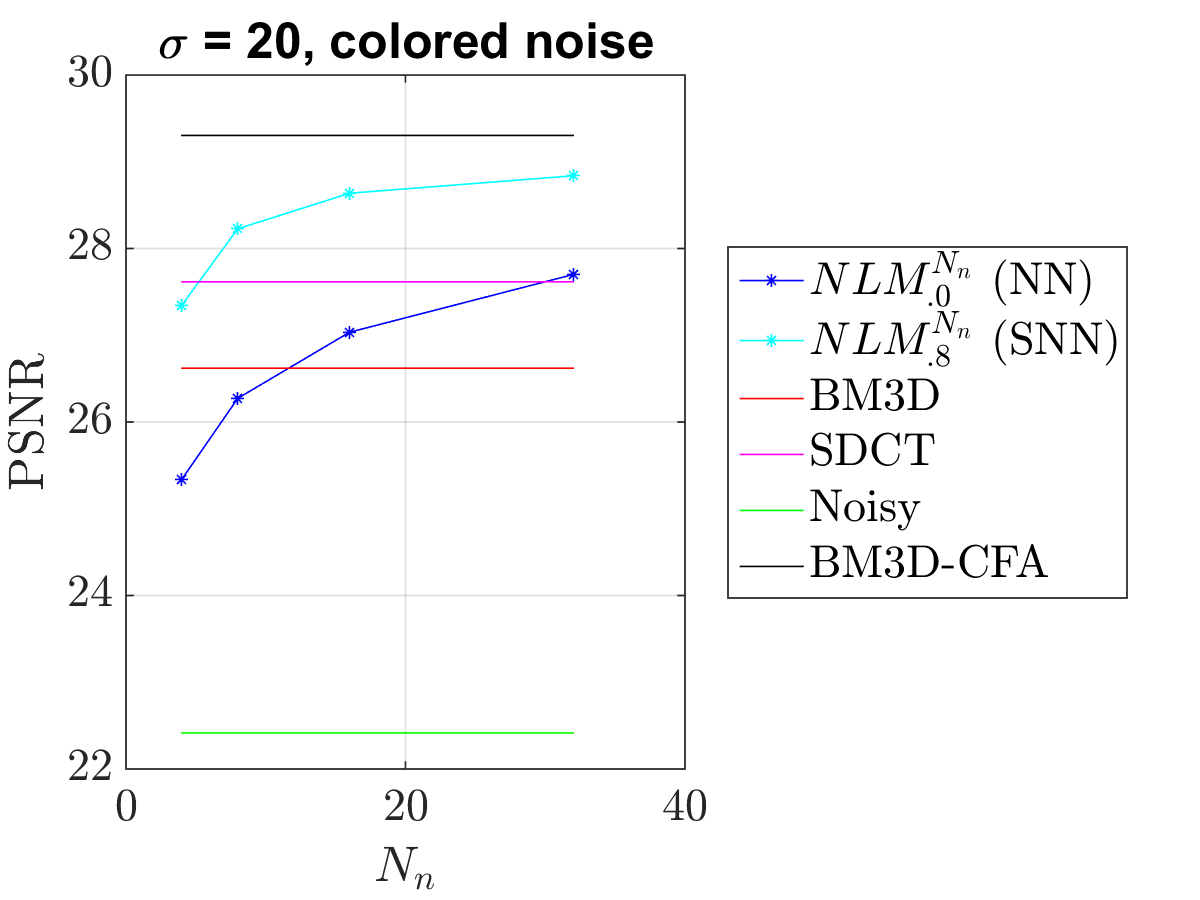}
\hspace{-2.0cm}
\includegraphics[width=0.24\textwidth]{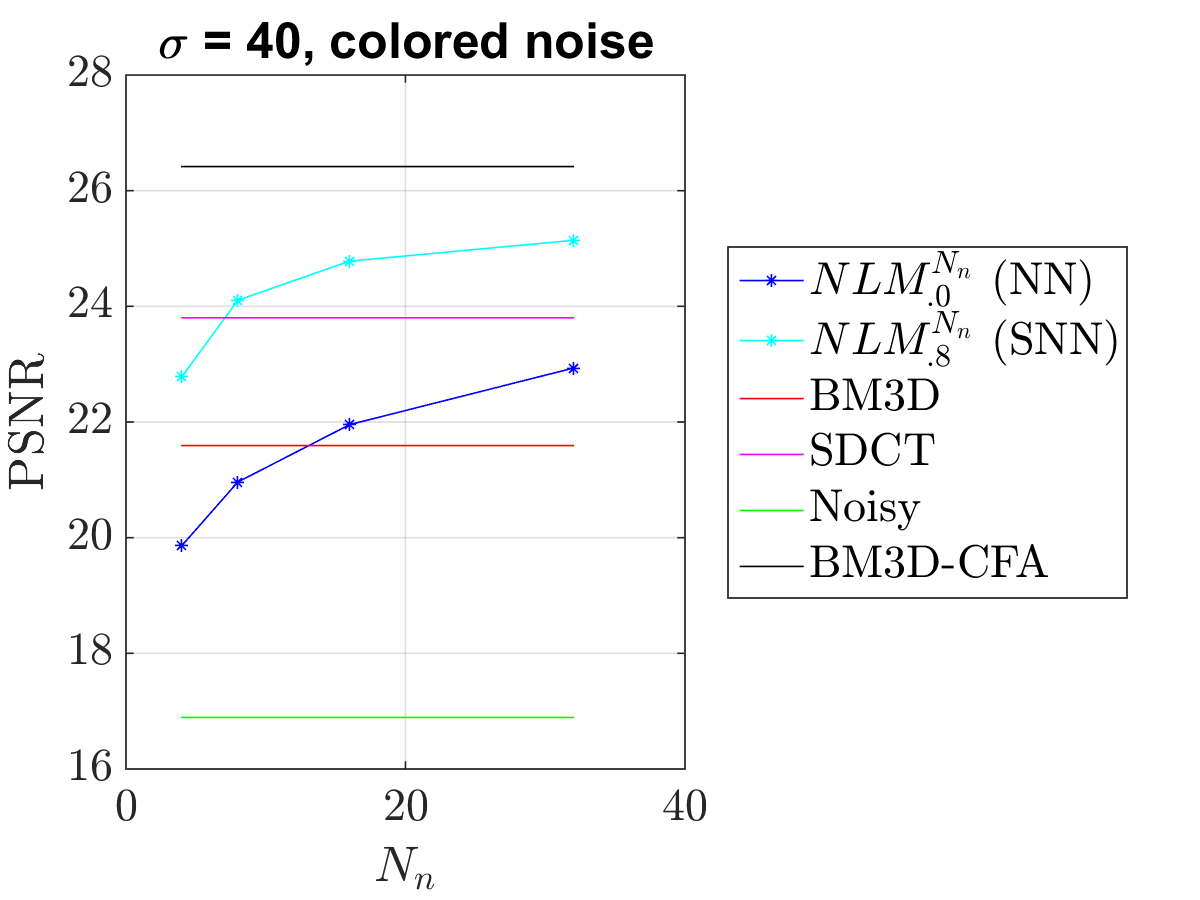}
\hspace{-2.0cm}
\includegraphics[width=0.24\textwidth]{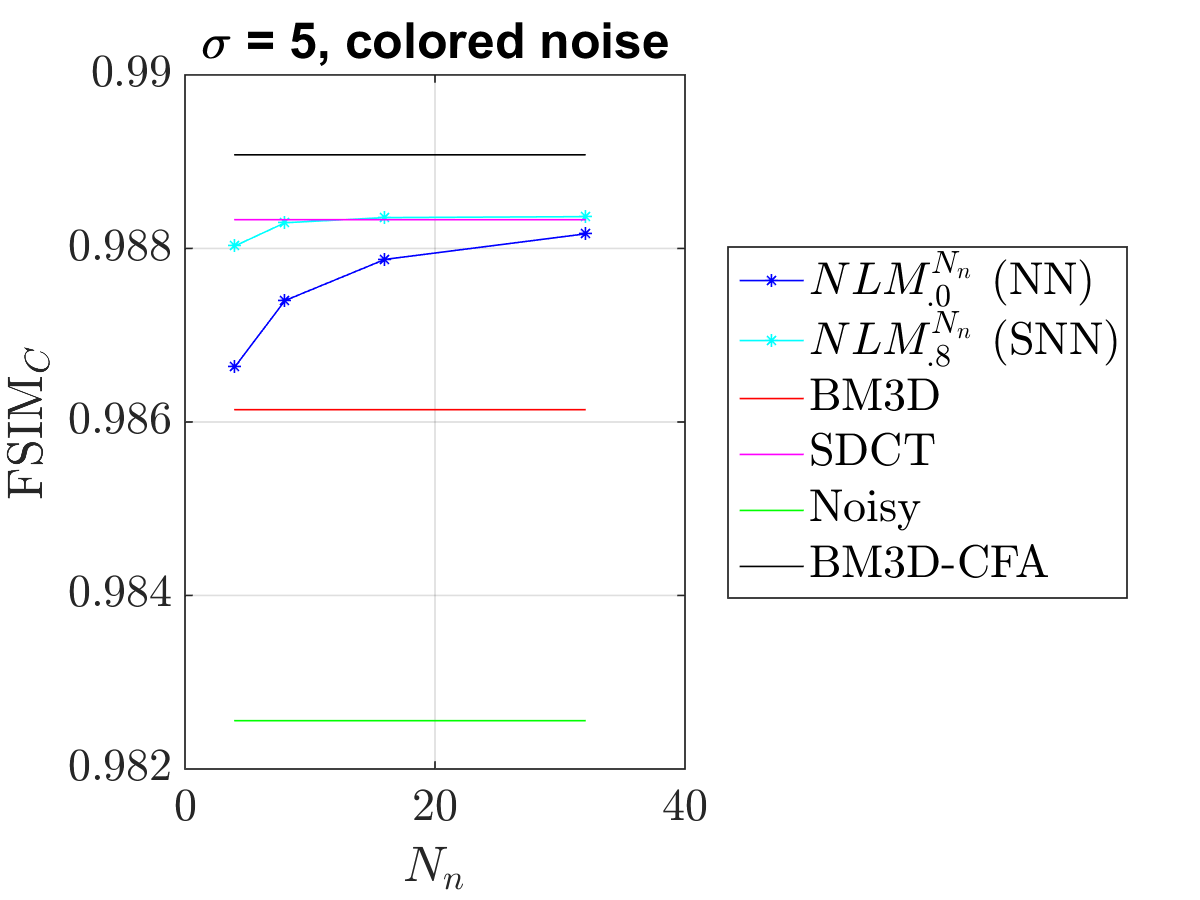}
\hspace{-2.0cm}
\includegraphics[width=0.24\textwidth]{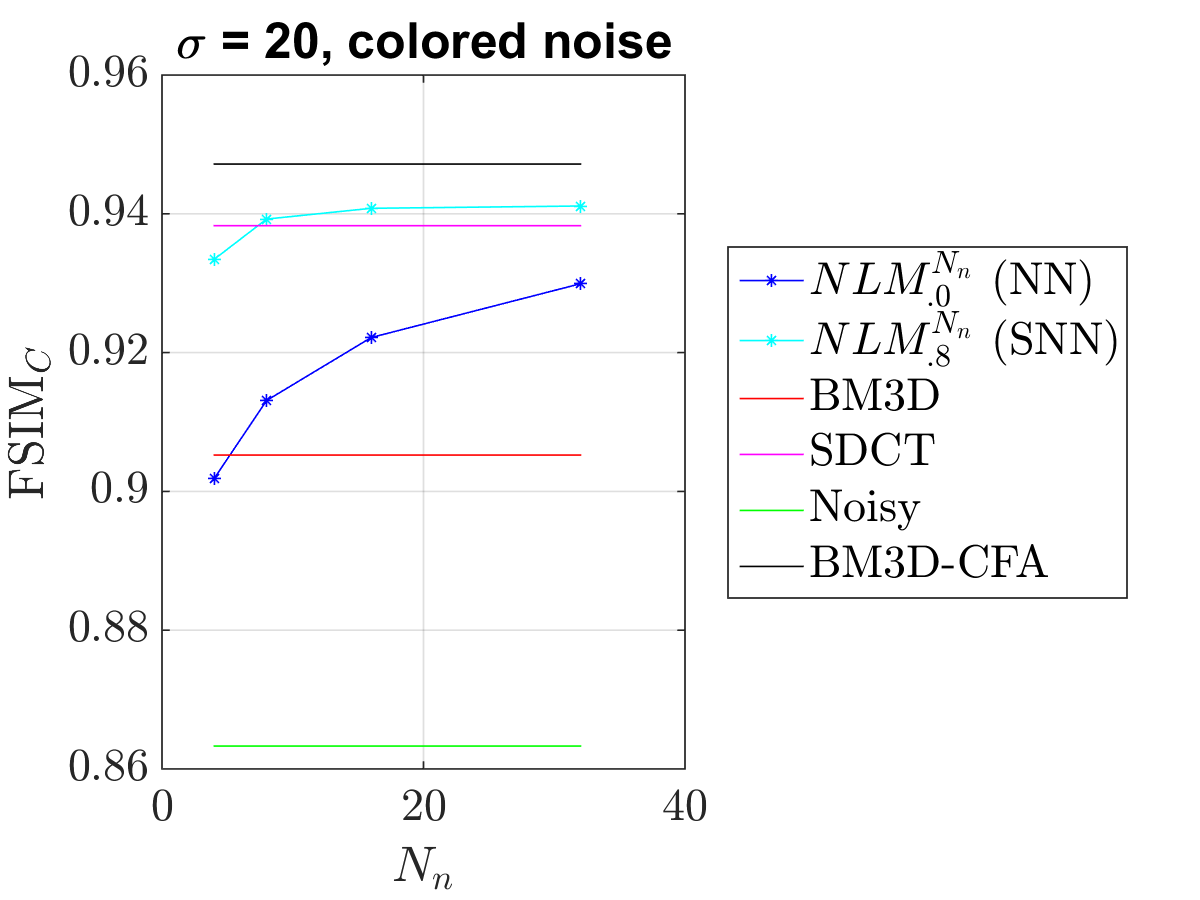}
\hspace{-2.0cm}
\includegraphics[width=0.24\textwidth]{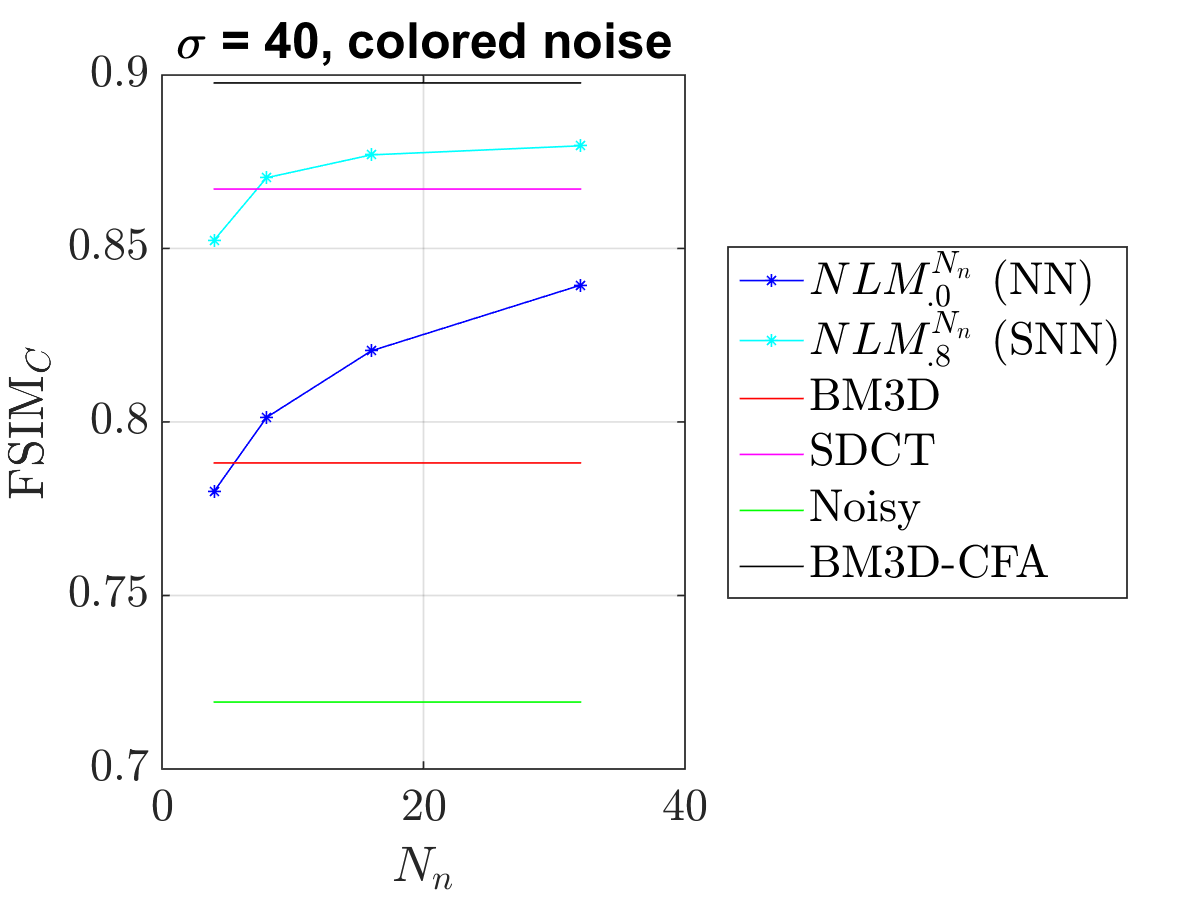}\\
\end{tabular}
\caption{Average PSNR and FSIM$_C$ measured on the Kodak dataset, as a function of the number of neighbors $N_n$, for NLM using the NN ($NLM^{N_n}_{.0}$) and SNN ($NLM^{N_n}_{.8}$) approaches, compared to SDCT~\cite{Yu11}, BM3D~\cite{Dab07}, and BM3D-CFA~\cite{Dan09} in the case of additive white Gaussian noise (first two rows) and colored noise (bottom rows). Notice that the number of patches used by BM3D and BM3D-CFA is constant and coherent with the optimal implementations described in~\cite{Dab07} and~\cite{Dan09}.}
\label{fig:scaling}
\end{figure*}

The first row of Fig.~\ref{fig:scaling} shows the average PSNR and FSIM$_C$ achieved by NLM using NN ($o=0$) and SNN ($o=0.8$) on the Kodak dataset, as a function of the number of neighbors $N_n$. Notice that the filter computational cost scales linearly (apart from the neighbors search part) with $N_n$. Compared to NN, the SNN approach always achieves an higher PSNR, since it smooths better the flat areas, as already shown in Fig.~\ref{fig:results_kodak} in the main paper. FSIM$_C$ is slightly lower for NN when $N_n \ge 16$, because SNN tends to oversmooth the edges, but when few neighbors are used the SNN approach clearly outperforms NN. The state-of-the-art denoising filter BM3D~\cite{Dab07} produces images of superior quality, but at a much higher computational cost. We also compared the image quality achieved by our approach with the Sliding Discrete Cosine Transform filter (SDCT~\cite{Yu11}), a denoising algorithm of comparable computational complexity based on signal sparsification and not using a NL approach. SDCT achieves a comparable PSNR, but generally a lower FSIM$_C$.

\subsection{Colored noise}

Although the case of white Gaussian noise is widely studied, it represents an ideal situation which rarely occurs in practice.
Having in mind the practical application of denoising, we also study the more general case of colored noise.
In fact, images are generally acquired using a Bayer sensor.
RGB data are obtained only after demosaicing, which introduces correlations between nearby pixels.
A direct consequence of this is that the NN approach is even more likely to collect false matches, amplifying low-frequency noise as well as demosaicing artifacts.
To study this, we mosaic the noisy images of the Kodak dataset, add noise in the Bayer domain, and subsequently demosaic them through the widely used Malvar algorithm~\cite{Mal04}.
Denoising is then performed through $NLM^{N_n}_{.0}$, $NLM^{N_n}_{.8}$ (for $N_n = \{4, 8, 16, 32\}$), BM3D~\cite{Dab07}, and SDCT~\cite{Yu11}. All these filters require an estimate of the noise standard deviation in the RGB domain,  obtained here considering that RGB data come from a linear combination (as in ~\cite{Mal04}) of independent samples in the Bayer domain.
To compare with the state-of-the-art, we also filter the images with BM3D-CFA~\cite{Dan09}, a modified version of BM3D specifically tailored to denoise data in the Bayer domain.  

The bottom rows of Fig.~\ref{fig:scaling} show the average PSNR and FSIM$_C$ on the Kodak dataset.
The proposed SNN approach is always more effective than NN in removing colored noise, both in terms of PSNR and FSIM$_C$.
This is clearly related with the higher occurrence of \emph{noise-to-noise} matching for NN in the case of colored noise. Visual inspection (Fig.~\ref{fig:coloredNoiseComparison} in the main paper) further confirms that the NN approach applied to NLM suffers from the \emph{noise-to-noise} matching problem.
NLM with SNN also generally outperforms SDCT and BM3D in terms of PSNR and FSIM$_C$. Since BM3D adopts a NN search selection strategy, it also suffers from the \emph{noise-to-noise} matching problem in this case.
Remarkably, for $\sigma \le 20$, NLM coupled with SNN is better, in terms of PSNR, than the state-of-the-art BM3D-CFA. It is slightly inferior for higher noise levels, but at a lower computational cost.
Visually, the result produced by NLM with SNN is comparable with that obtained by BM3D-CFA: our approach generates a slightly more noisy image, but with better preserved finer details (\emph{e.g.}, see the eyelids in Fig.~\ref{fig:coloredNoiseComparison} in the main paper) and without introducing grid artifacts.

\subsection{The case of a ``real'' image}

Having in mind the practical application of SNN, it is worthy analyzing the peculiarities of image denoising for ``real'' images.
We test different filtering techniques (SDCT, NLM$^{16}_{.0}$, NLM$^{16}_{.8}$, and BM3D-CFA) on a high resolution ($2592 \times 1944$) image, captured at ISO 1200 with an NVIDIA Shield tablet.
Notice that image resolution is much higher in this case, compared to the Kodak dataset.
Futhermore, the noise distribution in the Bayer domain is  far from the ideal, Gaussian distribution with constant variance.
Therefore, we compute the sensor noise model and the corresponding Variance Stabilizing Transform (VST) as in~\cite{Foi09} and apply the VST in the Bayer domain such that the noise distribution approximately resembles the ideal one.
We filter the image through the SDCT, NLM$^{16}_{.0}$, and NLM$^{16}_{.8}$ denoising algorithms in the RGB domain, after applying the VST and Malvar demosaicing, which introduces correlations among pixels.
The inverse VST is then applied to the data in the RGB domain.
Differently from the other algorithms, BM3D-CFA is applied directly to raw data, after the VST and before the inverse VST. This represents an advantage of BM3D-CFA over the other approaches, that have been designed to deal with white noise, but are applied in this case to colored noise.
Finally, after the denoising step, we apply color correction, white balancing, gamma correction, unsharp masking, coherently with the typical workflow of an Image Signal Processor; these steps can easily increase the visibility of  artifacts and residual noise. A stack of 650 frames is averaged to build a ground truth image of the same scene. 

\begin{table*}[h!]
\centering
\begin{tabular}{|r||c|c|c|c|c|}
\hline
& Noisy & SDCT & NLM$^{16}_{.0}$ (NN) & NLM$^{16}_{.8}$ (SNN) & BM3D-CFA\\
\hline
PSNR & 21.2689 & 23.3715 & 24.0972 & 24.5500   & 25.2580\\
SSIM & 0.8741 & 0.8889 & 0.9135 & 0.9261 &    0.9607\\
MSSSIM & 0.7133 & 0.7773 & 0.8124 & 0.8450 &   0.8943\\
GMSD & 0.1921 & 0.1499 & 0.1356 & 0.1145 &    0.0921\\
FSIM & 0.9920 & 0.9927 & 0.9933 & 0.9940 &    0.9952\\
FSIM$_C$ & 0.9890 & 0.9899 & 0.9911 &    0.9921 & 0.9941\\
\hline
\end{tabular}
\caption{Image quality metrics for the ``real'' image in Fig. \ref{fig:realImage}, acquired with and NVIDIA Shield Tablet at ISO 1200 and denoised with different algorithms.}
\label{tab:realImage}
\end{table*}

\begin{figure*}
\centering
\includegraphics[width=0.35\textwidth]{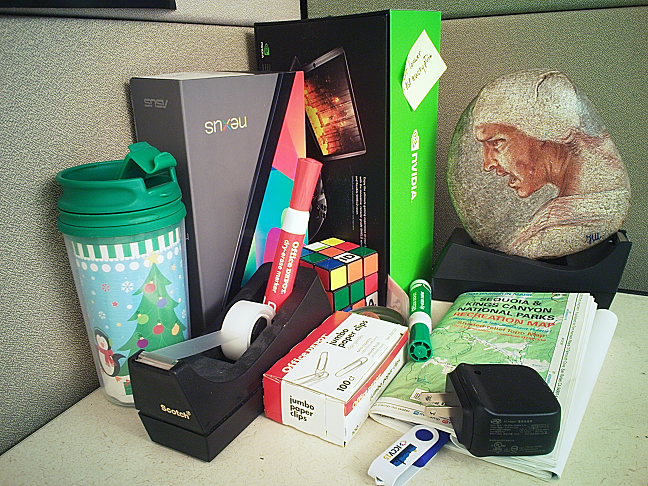}\\
\vspace{1mm}
\setlength\tabcolsep{0.05cm}
\begin{tabular}{ccccccc}
\rotatebox{90}{Ground truth}&
\includegraphics[width=0.14754\textwidth]{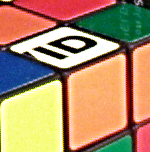}&
\includegraphics[width=0.14754\textwidth]{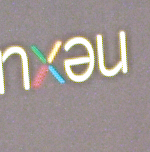}&
\includegraphics[width=0.14754\textwidth]{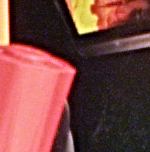}&
\includegraphics[width=0.14754\textwidth]{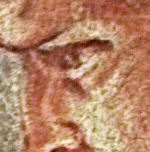}&
\includegraphics[width=0.14754\textwidth]{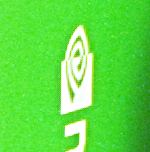}&
\includegraphics[width=0.14754\textwidth]{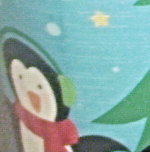}\\
\rotatebox{90}{Noisy}&
\includegraphics[width=0.14754\textwidth]{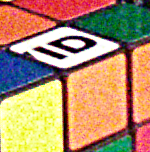}&
\includegraphics[width=0.14754\textwidth]{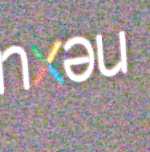}&
\includegraphics[width=0.14754\textwidth]{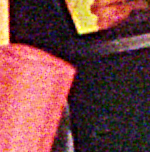}&
\includegraphics[width=0.14754\textwidth]{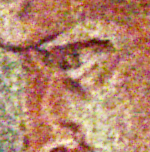}&
\includegraphics[width=0.14754\textwidth]{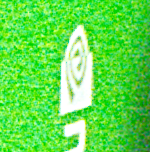}&
\includegraphics[width=0.14754\textwidth]{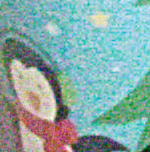}\\
\rotatebox{90}{SDCT}&
\includegraphics[width=0.14754\textwidth]{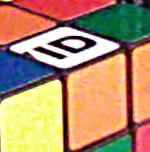}&
\includegraphics[width=0.14754\textwidth]{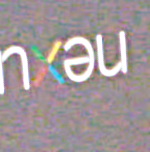}&
\includegraphics[width=0.14754\textwidth]{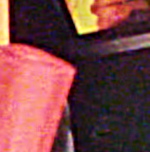}&
\includegraphics[width=0.14754\textwidth]{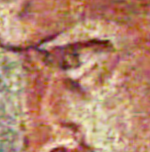}&
\includegraphics[width=0.14754\textwidth]{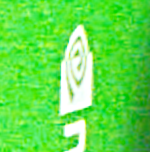}&
\includegraphics[width=0.14754\textwidth]{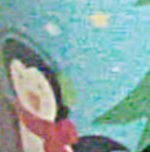}\\
\rotatebox{90}{NLM$^{16}_{.0}$ (NN)}&
\includegraphics[width=0.14754\textwidth]{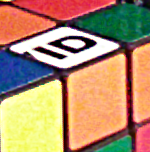}&
\includegraphics[width=0.14754\textwidth]{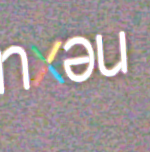}&
\includegraphics[width=0.14754\textwidth]{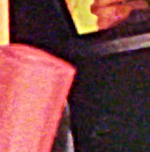}&
\includegraphics[width=0.14754\textwidth]{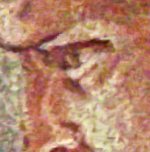}&
\includegraphics[width=0.14754\textwidth]{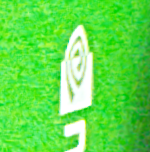}&
\includegraphics[width=0.14754\textwidth]{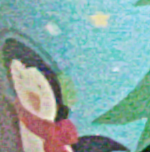}\\
\rotatebox{90}{NLM$^{16}_{.8}$ (SNN)}&
\includegraphics[width=0.14754\textwidth]{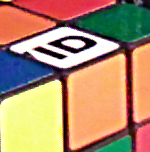}&
\includegraphics[width=0.14754\textwidth]{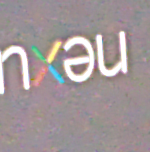}&
\includegraphics[width=0.14754\textwidth]{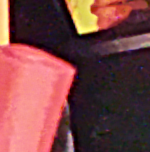}&
\includegraphics[width=0.14754\textwidth]{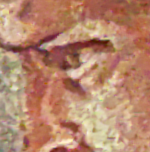}&
\includegraphics[width=0.14754\textwidth]{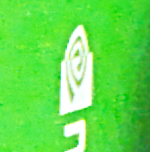}&
\includegraphics[width=0.14754\textwidth]{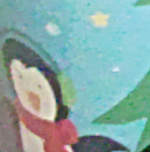}\\
\rotatebox{90}{BM3D-CFA}&
\includegraphics[width=0.14754\textwidth]{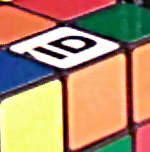}&
\includegraphics[width=0.14754\textwidth]{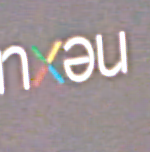}&
\includegraphics[width=0.14754\textwidth]{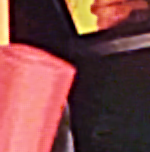}&
\includegraphics[width=0.14754\textwidth]{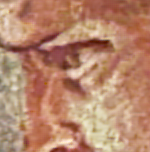}&
\includegraphics[width=0.14754\textwidth]{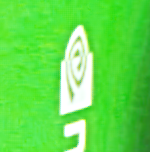}&
\includegraphics[width=0.14754\textwidth]{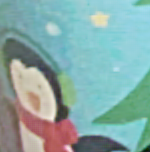}\\
\end{tabular}
\caption{Comparison of several patches from the image on the top, acquired with and NVIDIA Shield Tablet at ISO 1200, demosaiced and then denoised with the different algorithms. Color correction, white balancing, gamma correction and unsharp masking have also been applied after denoising. The ground truth image is obtained by averaging 650 frames. Better seen at 400\% zoom.}
\label{fig:realImage}
\end{figure*}

The results are shown in Fig.~\ref{fig:realImage}. Visual inspection confirms the superiority of NLM$^{16}_{.8}$ over SDCT and NLM$^{16}_{.0}$ even for the case of a real image. The NLM$^{16}_{.0}$ filter preserves high frequency details in the image, but it also generates high frequency colored noise in the flat areas, because of the \emph{noise-to-noise} matching problem. The SDCT filter produces a more blurry image with middle frequency colored noise in the flat areas. Coherently with the case of colored noise and the Kodak dataset, our approach provides an image of quality comparable to BM3D-CFA; the NLM$^{16}_{.8}$ filter is slightly more noisy than BM3D-CFA, as measured by the image quality metrics in Table \ref{tab:realImage}, but filtering requires a lower computational cost.

\bibliography{references}
\bibliographystyle{acm}

\end{document}

%% file: Introduction.tex

\section{Introduction}
\label{sec:Introduction}

\begin{figure*}
\centering
\setlength{\tabcolsep}{1pt} 
\def\arraystretch{0.85}
\begin{tabular}{cccccc}
Noisy & NLM$^{361}_{0.0}$ & NLM$^{16}_{0.0}$ & NLM$^{16}_{0.8}$ & NLM$^{16}_{1.0}$ & Ground truth\\
{\includegraphics[width=0.12\textwidth]{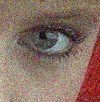}}&
{\includegraphics[width=0.12\textwidth]{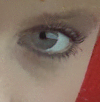}} &
{\includegraphics[width=0.12\textwidth]{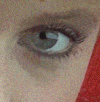}} &
{\includegraphics[width=0.12\textwidth]{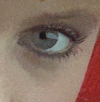}} &
{\includegraphics[width=0.12\textwidth]{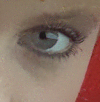}} &
{\includegraphics[width=0.12\textwidth]{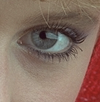}}
\vspace{-0.57cm}\\
\tiny \color{yellow} $ \begin{array}{c}
\bf{PSNR: 22.14} \\ \bf{FSIM_C: 0.7604}
\end{array} $
&
\tiny \color{yellow} $ \begin{array}{c}
\bf{PSNR: 32.10} \\ \bf{FSIM_C: 0.9153}
\end{array} $
&
\tiny \color{yellow} $ \begin{array}{c}
\bf{PSNR: 29.57} \\ \bf{FSIM_C: 0.9144}
\end{array} $
&
\tiny \color{yellow} $ \begin{array}{c}
\bf{PSNR: 31.01} \\ \bf{FSIM_C: 0.9255}
\end{array} $
&
\tiny \color{yellow} $ \begin{array}{c}
\bf{PSNR: 31.66} \\ \bf{FSIM_C: 0.9183}
\end{array} $
&
\vspace{0.12cm}\\ 
{\includegraphics[width=0.12\textwidth]{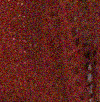}} &
{\includegraphics[width=0.12\textwidth]{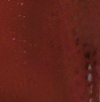}} &
{\includegraphics[width=0.12\textwidth]{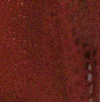}} &
{\includegraphics[width=0.12\textwidth]{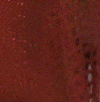}} &
{\includegraphics[width=0.12\textwidth]{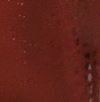}} &
{\includegraphics[width=0.12\textwidth]{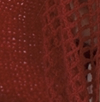}}
\vspace{-0.57cm}\\
\tiny \color{yellow} $ \begin{array}{c}
\bf{PSNR: 22.51} \\ \bf{FSIM_C: 0.7828}
\end{array} $
&
\tiny \color{yellow} $ \begin{array}{c}
\bf{PSNR: 31.92} \\ \bf{FSIM_C: 0.8294}
\end{array} $
&
\tiny \color{yellow} $ \begin{array}{c}
\bf{PSNR: 29.42} \\ \bf{FSIM_C: 0.8854}
\end{array} $
&
\tiny \color{yellow} $ \begin{array}{c}
\bf{PSNR: 31.04} \\ \bf{FSIM_C: 0.8778}
\end{array} $
&
\tiny \color{yellow} $ \begin{array}{c}
\bf{PSNR: 31.55} \\ \bf{FSIM_C: 0.8439}
\end{array} $
&
\vspace{0.12cm}\\
{\includegraphics[width=0.12\textwidth]{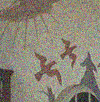}} &
{\includegraphics[width=0.12\textwidth]{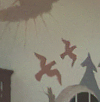}} &
{\includegraphics[width=0.12\textwidth]{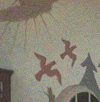}} &
{\includegraphics[width=0.12\textwidth]{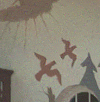}} &
{\includegraphics[width=0.12\textwidth]{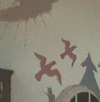}} &
{\includegraphics[width=0.12\textwidth]{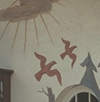}}
\vspace{-0.57cm}\\
\tiny \color{yellow} $ \begin{array}{c}
\bf{PSNR: 22.17} \\ \bf{FSIM_C: 0.7299}
\end{array} $
&
\tiny \color{yellow} $ \begin{array}{c}
\bf{PSNR: 33.16} \\ \bf{FSIM_C: 0.9412}
\end{array} $
&
\tiny \color{yellow} $ \begin{array}{c}
\bf{PSNR: 29.94} \\ \bf{FSIM_C: 0.9183}
\end{array} $
&
\tiny \color{yellow} $ \begin{array}{c}
\bf{PSNR: 31.78} \\ \bf{FSIM_C: 0.9364}
\end{array} $
&
\tiny \color{yellow} $ \begin{array}{c}
\bf{PSNR: 32.72} \\ \bf{FSIM_C: 0.9424}
\end{array} $
&
\vspace{0.12cm}\\
{\includegraphics[width=0.12\textwidth]{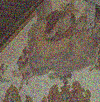}} &
{\includegraphics[width=0.12\textwidth]{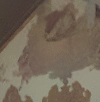}} &
{\includegraphics[width=0.12\textwidth]{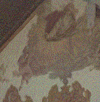}} &
{\includegraphics[width=0.12\textwidth]{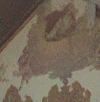}} &
{\includegraphics[width=0.12\textwidth]{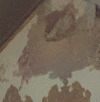}} &
{\includegraphics[width=0.12\textwidth]{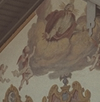}}
\vspace{-0.57cm}\\
\tiny \color{yellow} $ \begin{array}{c}
\bf{PSNR: 22.09} \\ \bf{FSIM_C: 0.7404}
\end{array} $
&
\tiny \color{yellow} $ \begin{array}{c}
\bf{PSNR: 31.41} \\ \bf{FSIM_C: 0.9107}
\end{array} $
&
\tiny \color{yellow} $ \begin{array}{c}
\bf{PSNR: 29.52} \\ \bf{FSIM_C: 0.9124}
\end{array} $
&
\tiny \color{yellow} $ \begin{array}{c}
\bf{PSNR: 30.70} \\ \bf{FSIM_C: 0.9249}
\end{array} $
&
\tiny \color{yellow} $ \begin{array}{c}
\bf{PSNR: 31.11} \\ \bf{FSIM_C: 0.9151}
\end{array} $
&
\end{tabular}
\caption{Noisy, $100 \times 100$ patches from the Kodak dataset, corrupted by zero-mean Gaussian noise, $\sigma = 20$. Traditional NLM (NLM$^{361}_{0.0}$) uses 361, $3 \times 3$ patches in a $21 \times 21$ search window; it denoises effectively the flat areas (the skin, the wall), but it blurs the small details (texture of the textile, fresco details). Using 16 NNs for each patch (NLM$^{16}_{0.0}$) improves the small details, but introduces colored noise in the flat areas. The proposed SNN technique (NLM$^{16}_{1.0}$) uses 16 neighbors and mimics the results of traditional NLM in flat areas (high PSNR), while keeping the visibility of small details (high FSIM$_C$). Best results are achieved when patches are collected through SNN, with $o = 0.8$ (NLM$^{16}_{0.8}$). Better seen at 400\% zoom.}
\label{fig:results_kodak}
\vspace{-1em}
\end{figure*}

Non-Local-Means (NLM) denoising has been widely investigated by researchers\cite{Bua05,Leb12,Bua11}.
Denoising of a given patch is obtained as a weighted average of the surrounding patches, with weights proportional to the patch similarity.
The filtering parameters such as the number and size of the patches affect both the quality of the output images \cite{Duv11}, and the computational burden of the filter\cite{Tsa15}.
A widely used practice is to reduce the number of neighbors collected for each reference patch:  the 3D Block-Matching (BM3D) filter achieves state-of-the-art results in this way~\cite{Dab07}.
The neighbors' set is collected through a Nearest-Neighbors (NN) approach.
Reducing the set size leads to images with sharp edges \cite{Duv11}, but it also introduces low-frequency artifacts (Fig.~\ref{fig:results_kodak}).

Our contribution is to show that  this artifact occurs because the estimate of the noise-free patch from the set of NNs is biased.
The the best of our knowledge, this is the first time this problem is explicitely investigated, although other authors (e.g.~\cite{Leb12, Duv11, Wu13}) analyzed other sources of bias in NLM.
We propose a strategy to collect neighbors, named Statistical NN (SNN), which reduces the prediction error of the estimate of the noise-free patch.
Using fewer neighbors, SNN leads to an improvement in perceived image quality, both in case of white and colored Gaussian noise. In the latter case, visual inspection reveals that NLM with SNN achieves an image quality comparable to the state-of-the-art, at a much lower computational cost.

%% file: RelatedWork.tex

\section{Related work}
\label{sec:RelatedWork}

NLM denoising~\cite{Bua05, Bua11} averages similar patches and then aggregates the averages. Given a noisy reference patch with $P$ elements, $\boldsymbol{\mu_r} = [\mu_r^0 \ \mu_r^1.... \ \mu_r^{P-1}]$, its squared distance from a noisy neighbor patch $\boldsymbol{\gamma_k}$, is $
\delta^2 \left(\boldsymbol{\mu_r}, \boldsymbol{\gamma_k} \right)
=
\frac{1}{P} \cdot
\sum_{i = 0}^{P-1}{\left(\mu_r^i - \gamma_k^i \right)^2}$. Following~\cite{Bua11}, the weight of $\boldsymbol{\gamma_k}$ in the average is:
\begin{equation}
w_{\boldsymbol{\mu_r}, \boldsymbol{\gamma_k}} = \text{exp}\left\{\text{max}[0, \delta^2 \left(\boldsymbol{\mu_r}, \boldsymbol{\gamma_k} \right) - 2 \sigma^2]\right\} / {h^2},
\label{eq:weight}
\end{equation}
where $h$ is the \emph{filtering parameter} and $\sigma^2$ is the variance of zero-mean, white Gaussian noise~\cite{Bua05, Bua11}.
The estimate of the noise-free patch, $\hat{\boldsymbol{\mu}}({\boldsymbol{\mu_r}})$, is then: 
\begin{equation}
\hat{\boldsymbol{\mu}}({\boldsymbol{\mu_r}}) = \sum_k w_{\boldsymbol{\mu_r}, \boldsymbol{\gamma}_k} \cdot \boldsymbol{\gamma}_k  \;\;\; / \;\; \sum_k w_{\boldsymbol{\mu_r}, \boldsymbol{\gamma}_k}.
\label{eq:denoise}
\end{equation}
Using multiple scales in the distance function can improve patch matching~\cite{Lou09,Lot2016}, leading to higher quality of the filtered images at the price of a higher computational cost.
Wu et al.~\cite{Wu13} identify a source of bias in $\hat{\boldsymbol{\mu}}(\boldsymbol{\mu_r})$, due to the correlation among partially overlapping patches, and propose a weighting scheme to take this into account. Duval et al.~\cite{Duv11} interpret NLM denoising as a bias~/~variance dilemma and show that $\hat{\boldsymbol{\mu}}(\boldsymbol{\mu_r})$ is biased even in absence of noise; reducing the number of neighbors decreases the cost of the filter and the bias at the same time, but it also increases the variance on $\hat{\boldsymbol{\mu}}(\boldsymbol{\mu_r})$,  leaving some residual noise in the image.
%
%
State-of-the art algorithms like BM3D~\cite{Dab07}, BM3D-SAPCA~\cite{Dab09}, and NL-Bayes~\cite{Leb12} use a reduced set of neighbors, but they also require additional, computationally intensive processing steps and continue to suffer from visible artifacts on sharp edges  and in smooth regions~\cite{Kna14}. Machine learning can be used to learn to combine a set of  patches for denoising, but the pros and cons of using a set of NN patches have not been discussed~\cite{Ahn17}.

%% file: Method.tex
\section{Method}
\label{sec:Method}

None of the existing methods focus their attention on the fact that the neighbors' selection criterion, combined with the fact that the reference patch $\boldsymbol{\mu_r}$ is noisy, affects the prediction error of $\hat{\boldsymbol{\mu}}(\boldsymbol{\mu_r})$. In the \emph{Additional Material}, we resort to a toy problem to compute the bias in $\hat{\boldsymbol{\mu}}(\boldsymbol{\mu_r})$ introduced by the NN search strategy and we show analytically that collecting neighbors with the SNN approach can dramatically reduce it. For reasons of space, we describe here only the guiding principle behind SNN, and invite the curious reader to read the \emph{Additional Material} for the math details.

\begin{wrapfigure}{l}{0.54\textwidth}
\begin{center}
\includegraphics[width=0.26\textwidth,clip=true,trim={0cm 0cm 14cm 0cm}]{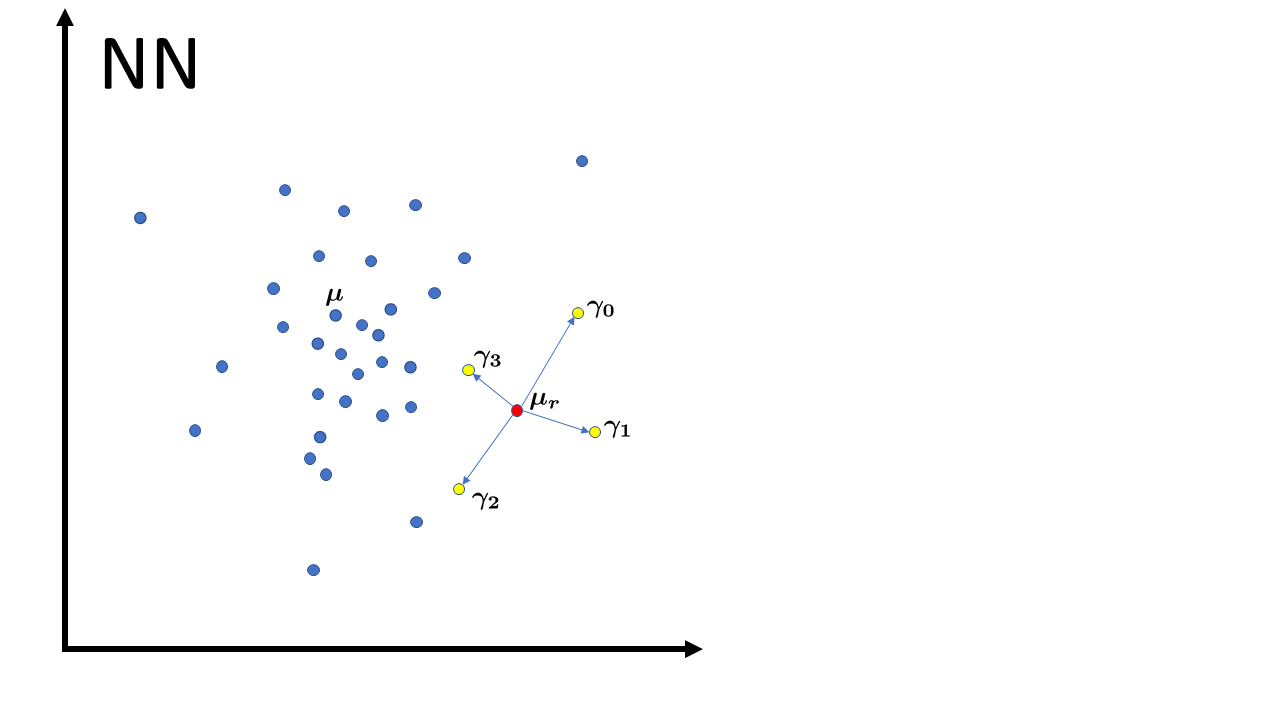}
\includegraphics[width=0.26\textwidth,clip=true,trim={0cm 0cm 14cm 0cm}]{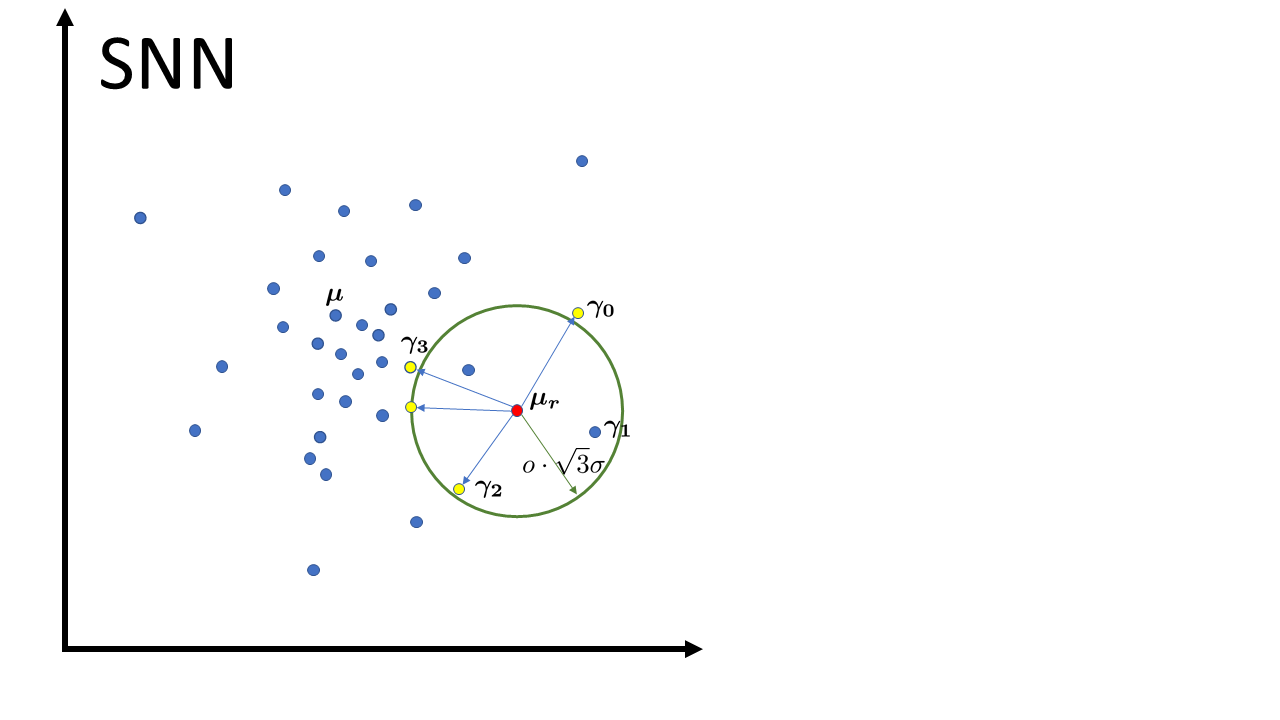}
\end{center}
\caption{Left: collecting neighbors $\{\boldsymbol{\gamma_k}\}_{k=0..N-1}$ around the reference, noisy patch $\boldsymbol{\mu_r}$ through the traditional NN approach generates a neighbor set which is highly biased towards $\boldsymbol{\mu_r}$. The proposed SNN approach (right) collects neighbors close to the distance $o \cdot \sqrt(3) \sigma$ from $\boldsymbol{\mu_r}$, where $\sigma$ is the noise standard deviation and $o$ and additional \emph{offset} parameter. The resulting set $\{\boldsymbol{\gamma_k}\}_{k=0..N-1}$ is less biased towards $\boldsymbol{\mu_r}$ and closer to the noise-free value $\boldsymbol{\mu}$.}
\label{fig:NNvsSNN}
\end{wrapfigure}

Let's assume that a set of $N$ neighbors, $\{\boldsymbol{\gamma_k}\}_{k=0..N-1}$, has to be collected for a noisy, reference patch $\boldsymbol{\mu_r}$. Fig. \ref{fig:NNvsSNN} shows the case of a two-dimensional patch. When $N$ neighbors are collected through the NN approach (left panel), the average of the set of neighbors is clearly biased toward $\boldsymbol{\mu_r}$. In NLM denoising, this drawback shows up as the \emph{noise-to-noise} matching issue, \emph{i.e.}, residual noise correlated with the reference patch is still present in the filtered image.

The recipe for collecting neighbors through the SNN approach is different and illustrated in the right panel. First of all, we notice that, in presence of white, Gaussian noise with standard deviation $\sigma$, the expected squared distance between the noisy, reference patch and a noisy neighbor patch is $E[\delta^2\left(\boldsymbol{\mu_r}, \boldsymbol{\gamma_k} \right)] = 2 \sigma^2$. The SNN neighbors are then defined as the patches $\{\boldsymbol{\gamma}_k\}_{k=1.. N_n}$ that minimize:
\begin{equation}
|\delta^2 \left(\boldsymbol{\mu_r}, \boldsymbol{\gamma}_k \right) - o \cdot 2\sigma^2|,
\label{eq:SNNs}
\end{equation}
where where the \emph{offset} parameter $o$ allows to continuously move from the traditional NN approach ($o=0$) to the SNN approach ($o=1$). The right panel in Fig. \ref{fig:NNvsSNN} shows that the average of the set of SNNs is more likely to lie close to the noise-free patch $\boldsymbol{\mu}$. In fact, because of the statistical distribution of the noisy patches, the chance to collect a neighbor $\boldsymbol{\gamma_k}$ on the left side of $\boldsymbol{\mu_r}$ (and thus close to $\boldsymbol{\mu_r}$) is high, while neighbors on the right side of $\boldsymbol{\mu_r}$ are unlikely to occur. In practice, the SNN criterion looks for similar patches having orthogonal realizations of the noise.
This minimizes the \emph{noise-to-noise} matching issue and it leads to more effective noise cancellation, at the price of a slightly higher variance of $\hat{\mu}(\mu_r)$ (due to the fact that the neighbors are generally farther from $\boldsymbol{\mu_r}$).

%% file: Results.tex

\section{Results}
\label{sec:Results}

For reasons of space, a detailed numerical evaluation of the effectiveness of the SNN schema for NLM denoising, obtained thorugh a wide set of image quality metric (PSNR, SSIM\cite{Wan04}, MSSSIM\cite{Wan03}, GMSD\cite{Xue14}, FSIM, and FSIM$_C$\cite{Zha11b}), is reported only in the \emph{Additional Material}. It is important noticing that these numerical evaluation confirms the superiority of the SNN approach, but we discuss here only the qualitative results.

The case of white, Gaussian noise is illustrated in Fig. \ref{fig:results_kodak}. This shows a higher level of details when reducing the number of NN neighbors (from traditional NLM with 361 patches, NLM$^{361}_{0.0}$, to NLM with only 16 patches, NLM$^{16}_{.0}$), coherently with the reduction of the NLM bias for a small set of neighbors, already explained in~\cite{Duv11}. At the same time, the PSNR is decreasing, mostly because of the residual noise left in the flat areas (the skin of the girl, the wall with the fresco in Fig.~\ref{fig:results_kodak}); this is how the $noise-to-noise$ matching problem manifests in the NLM filtered image. Increasing the offset from $o = 0.0$ (NN) to $o = 1.0$ (SNN) increases the PSNR: SNN removes more noise in the flat areas (\emph{e.g.}, the girl's skin, the wall surface) compared to NN, thanks to the reduction of the bias introduced by the \emph{noise-to-noise} matching issue. On the other hand, low-contrast edges (\emph{e.g.}, the texture of the textile, the sun rays and the small details in the fresco) tend to be blurred by SNN. The best compromise between preserving low-contrasted details in the image and effectively smoothing the flat areas is obtained for $o = 0.8$, as shown in Fig. \ref{fig:results_kodak}.

\begin{figure*}
\centering
\setlength\tabcolsep{0.2cm}
\begin{tabular}{cccc}
Ground truth & BM3D-CFA~\cite{Dan09} & $NLM^{32}_{.0}$ (NN) & $NLM^{32}_{.8}$ (SNN)\\ 
\begin{overpic}[width=0.22\textwidth]{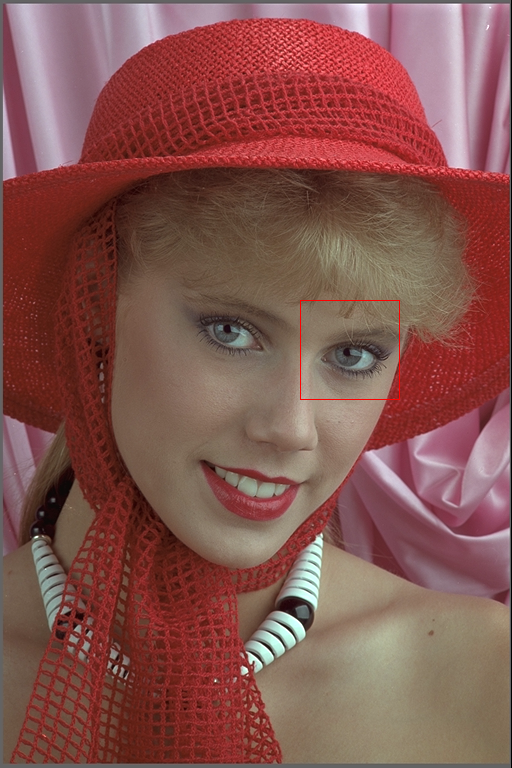}
 \put (-3,-3) {\includegraphics[width=0.12\textwidth]{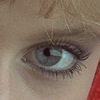}}
\end{overpic}
&
\begin{overpic}[width=0.22\textwidth]{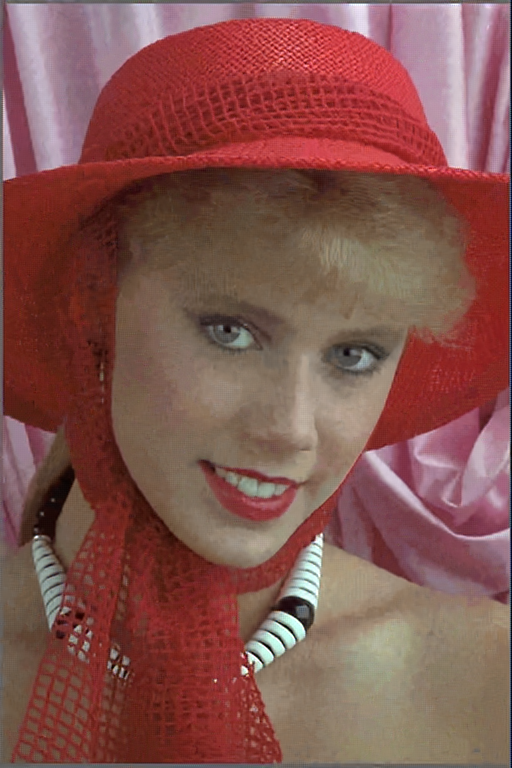}
 \put (-3,-3) {\includegraphics[width=0.12\textwidth]{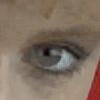}}
\end{overpic}
&
\begin{overpic}[width=0.22\textwidth]{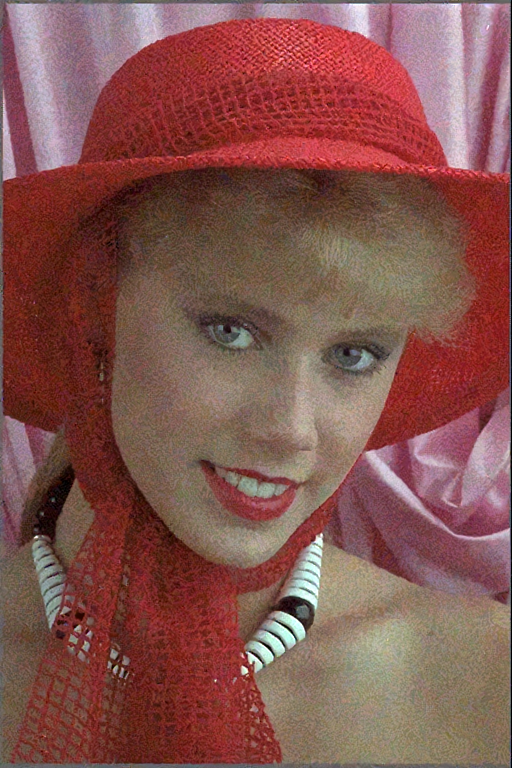}
 \put (-3,-3) {\includegraphics[width=0.12\textwidth]{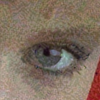}}
\end{overpic}
&
\begin{overpic}[width=0.22\textwidth]{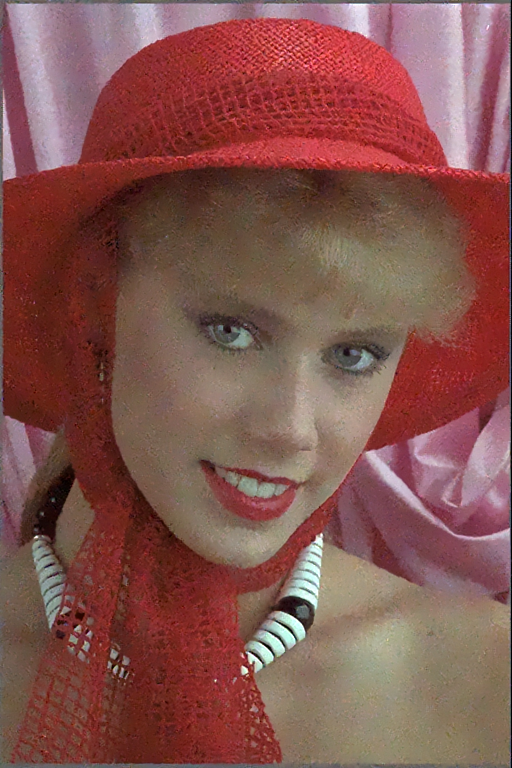}
 \put (-3,-3) {\includegraphics[width=0.12\textwidth]{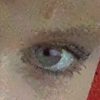}}
\end{overpic}
\end{tabular}
\caption{Comparison of different algorithms for colored noise denoising, $\sigma = 20$. The state-of-the-art BM3D-CFA~\cite{Dan09} achieves the highest image quality metrics (see Fig.~\ref{fig:scaling} in the Additional Material); a grid artifact is however visible in the white part of the eye. The traditional NLM algorithm, using $N_n = 32$ NNs ($NLM^{32}_{.0}$), does not eliminate the colored noise introduced by demosaicing. Instead, when SNNs neighbors are used ($NLM^{32}_{.8}$), the colored noise is efficiently removed and quality at visual inspection is comparable to that of BM3D-CFA. In fact, our result appears less ``splotchy" and a bit sharper. Better seen at  400\% zoom.}
\label{fig:coloredNoiseComparison}
\vspace{-1em}
\end{figure*}

Having in mind denoising in practical situations, we also study the case of colored noise.
In fact, images are generally acquired using a Bayer sensor.
RGB data are obtained only after demosaicing, which introduces correlations between nearby pixels.
In this case the NN approach is even more likely to collect false matches, amplifying low-frequency noise and demosaicing artifacts.
Fig.~\ref{fig:coloredNoiseComparison} shows the comparison of state-of-the-art BM3D-CFA~\cite{Dan09} to NLM with a NN or SNN approach in the case of an image corrupted by Gaussian noise ($\sigma=20$) in the Bayer domain and then demosaiced. 
Visual inspection (Fig.~\ref{fig:coloredNoiseComparison}) confirms that the NLM with NN continues to suffer from the \emph{noise-to-noise} matching problem in the case of colored noise.
NLM with SNN is comparable in terms of image quality to BM3D-CFA: our approach generates a slightly more noisy image, but with better preserved finer details (\emph{e.g.}, see the eyelids in Fig.~\ref{fig:coloredNoiseComparison}) and without introducing grid artifacts that are on the other hand visible for BM3D-CFA. Overall, our approach and BM3D-CFA result to be comparable in terms of image quality (as further demonstrated numerically in the Additional Material), but NLM with SNN has a much lower computational cost.

Further numerical comparison against against SDCT~\cite{Yu11}, BM3D~\cite{Dab07} (in the case of white and colored noise), and BM3D-CFA~\cite{Dan09} (in the case of colored noise) are reported in the \emph{Additional Material}, where we also report and discuss the case of a ``real'' image.

%% file: Discussion.tex

\section{Discussion}
\label{sec:Discussion}


The NLM algorithm has been be analyzed as a bias / variance dilemma: Duval et al.~\cite{Duv11} showed that a reduction of the number of neighbors, reduces the bias in the estimate $\hat{\boldsymbol{\mu}}_{\boldsymbol{r}}$. The price to be paid is the introduction of the \emph{noise-to-noise} matching issue, that shows up as splotchy, colored artifacts in the filtered images. In fact, since the NNs of any noisy reference patch $\boldsymbol{\mu_r}$ lie close to it, their average is biased towards $\boldsymbol{\mu_r}$ itself. We have shown here that this new source of bias is associated to the strategy adopted to collect the set of neighbors and demonstrated that the SNN approach can largely mitigate this bias both in theory and in practice. 

When applied to NLM denoising and compared to the traditional NN strategy, the SNN approach produces images of quality similar to the original NLM in the flat areas, while keeping the visiblity of the small details.
In other words, SNN looks for neighbors whose noise realization is likely to be orthogonal to that of the reference patch.
Averaging these patches is consequently more likely to effectively cancel out the noise. 

The advantage of using the SNNs is even more evident in the case of colored noise, when \emph{noise-to-noise} matching is more likely to occur. Visual inspection reveals that in this case NLM with SNN achieves an image quality comparable to the state-of-the-art BM3D-CFA~\cite{Dan09}, but at a lower computational cost (see Fig.~\ref{fig:coloredNoiseComparison}). This case is indeed of practical importance, since denoising is performed after demosaicing in ``real'' images. Our experiments (see the \emph{Additional Material}) suggest that NLM with SNN is close to BM3D-CFA also in the case of ``real'' images, where white balance, color correction, gamma correction and edge enhancement are applied after denoising, and they can significantly  enhance the visibility of artifacts and residual noise.

It is worth mentioning that both the NN and SNN approaches to NLM denoising require the knowledge of the noise power, $\sigma^2$, to properly set the filtering parameter $h$ in Eq. (\ref{eq:weight}) and guide the selection of the neighbors (in the case of SNN). In the case of ``real'' images, several effective noise estimation methods have been proposed~\cite{Leb12}, but it is also worthy mentioning that the noise distribution in real images is far more complicated than zero-mean Gaussian~\cite{Foi09}. Even if a comprehensive analysis of the application of SNN to NLM denoising for real images goes beyond the scope of this paper, our approach can be effectively applied to ``real'' images through the application of a VST~\cite{Foi09}, and it achieves image quality far superior compared to traditional NLM and in practice comparable at visual inspection to the state-of-the-art BM3D-CFA approach~\cite{Dan09}, at a lower computational cost.

%% file: Tables/NLM_sigmaall.tex

\begin{table}
\renewcommand{\arraystretch}{1.25}
\begin{center}
\resizebox{0.75\linewidth}{!}{%
\begin{tabular}{|c|c|c|c|c|c|c|c|}
\hline
 & PSNR & SSIM & MSSSIM & GMSD & FSIM & FSIM$_C$ & $\sigma$\\ \hline 
noisy & 34.15 & .9625 & .9797 & .0211 & .9937 & .9931 & \multirow{11}{*}{5}\\
\cline{1-7}
NLM$^{361}_{.0}$ & \bf{38.34} & .9851 & .9921 & .0130 & .9941 & .9939 & \\
NLM$^{16}_{.0}$ & 38.08 & .9847 & .9919 & .\bf{0113} & \bf{.9949} & \bf{.9947} & \\
\cline{1-7}
NLM$^{16}_{.1}$ & 38.08 & .9847 & .9919 & \bf{.0113} & \bf{.9949} & \bf{.9947} & \\
NLM$^{16}_{.35}$ & 38.08 & .9847 & .9919 & \bf{.0113} & \bf{.9949} & \bf{.9947}& \\
NLM$^{16}_{.65}$ & 38.14 & .9850 & .9920 & \bf{.0113} & \bf{.9949} & \bf{.9947} &\\
NLM$^{16}_{.8}$ & 38.23 & \bf{.9854} & .9922 & .0115 & .9948 & .9946 & \\
NLM$^{16}_{.9}$ & 38.28 & \bf{.9854} & \bf{.9923} & .0118 & .9946 & .9945 & \\
NLM$^{16}_{1.0}$ & 38.29 & .9852 & .9921 & .0124 & .9944 & .9942 &\\
\hline 
%
noisy & 28.13 & .8744 & .9332 & .0667 & .9763 & .9743 & \multirow{11}{*}{10}\\
\cline{1-7}
NLM$^{361}_{.0}$ & \bf{34.84} & .9634 & .9804 & .0338 & .9814 & .9811 & \\
NLM$^{16}_{.0}$ & 33.96 & .9580 & .9773 & .0269 & \bf{.9861} & .9856 &\\
\cline{1-7}
NLM$^{16}_{.1}$ & 33.96 & .9580 & .9773 & .0269 & \bf{.9861} & .9856 & \\
NLM$^{16}_{.35}$ & 33.96 & .9580 & .9773 & .0269 & \bf{.9861} & .9856 & \\
NLM$^{16}_{.65}$ & 34.19 & .9610 & .9789 & \bf{.0266} & \bf{.9861} & \bf{.9857} & \\
NLM$^{16}_{.8}$ & 34.51 & .9639 & .9805 & .0274 & .9856 & .9852 & \\
NLM$^{16}_{.9}$ & 34.66 & \bf{.9645} & \bf{.9809} & .0291 & .9845 & .9841 & \\
NLM$^{16}_{1.0}$ & 34.71 & .9635 & .9803 & .0319 & .9828 & .9824 & \\
\hline
%
noisy & 22.11 & .6810 & .8236 & .1549 & .9250 & .9174 & \multirow{11}{*}{20}\\
\cline{1-7}
NLM$^{361}_{.0}$ & \bf{31.18} & .9109 & .9505 & .0749 & .9475 & .9468 & \\
NLM$^{16}_{.0}$ & 29.21 & .8802 & .9343 & .0607 & .9646 & .9633 & \\
\cline{1-7}
NLM$^{16}_{.1}$ & 29.21 & .8802 & .9343 & .0607 & .9646 & .9633 & \\
NLM$^{16}_{.35}$ & 29.21 & .8802 & .9343 & .0607 & .9646 & .9633 & \\
NLM$^{16}_{.65}$ & 29.75 & .8948 & .9420 & \bf{.0572} & \bf{.9653} & \bf{.9641} & \\
NLM$^{16}_{.8}$ & 30.45 & .9086 & .9493 & .0583 & .9631 & .9621 & \\
NLM$^{16}_{.9}$ & 30.81 & \bf{.9119} & \bf{.9511} & .0635 & .9585 & .9576 &\\
NLM$^{16}_{1.0}$ & 30.93 & .9084 & .9490 & .0719 & .9510 & .9503 &\\
\hline
noisy & 18.59 & .5331 & .7243 & .2138 & .8687 & .8536 & \multirow{11}{*}{30}\\
\cline{1-7}
NLM$^{900}_{.0}$ & \bf{29.11} & .8631 & .9197 & .1071 & .9154 & .9144 & \\
NLM$^{16}_{.0}$ & 27.40 & .8273 & .9035 & .0827 & .9438 & .9419 &\\
\cline{1-7}
NLM$^{16}_{.1}$ & 27.40 & .8273 & .9035 & .0827 & .9438 & .9419 &\\
NLM$^{16}_{.35}$ & 27.40 & .8273 & .9035 & .0827 & .9438 & .9419 &\\
NLM$^{16}_{.65}$ & 27.48 & .8311 & .9055 & .0817 & \bf{.9441} & \bf{.9422}& \\
NLM$^{16}_{.8}$ & 28.15 & .8543 & .9175 & \bf{.0793} & .9439 & \bf{.9422}&\\
NLM$^{16}_{.9}$ & 28.69 & \bf{.8665} & \bf{.9233} & .0850 & .9382 & .9368&\\
NLM$^{16}_{1.0}$ & 28.91 & .8640 & .9201 & .0988 & .9247 & .9235&\\
\hline
noisy & 16.09 & .4280 & .6414 & .2501 & .8157 & .7937 & \multirow{11}{*}{40}\\
\cline{1-7}
NLM$^{900}_{0.0}$ & \bf{27.71} & .8184 & .8893 & .1303 & .8856 & .8844&\\
NLM$^{16}_{0.0}$ & 25.38 & .7522 & .8591 & .1061 & .9237 & .9206&\\
\cline{1-7}
NLM$^{16}_{0.1}$ & 25.38 & .7522 & .8591 & .1061 & .9237 & .9206&\\
NLM$^{16}_{0.2}$ & 25.38 & .7522 & .8591 & .1061 & .9237 & .9206&\\
NLM$^{16}_{0.35}$ & 25.38 & .7522 & .8591 & .1061 & .9237 & .9206&\\
NLM$^{16}_{0.5}$ & 25.38 & .7522 & .8591 & .1061 & .9237 & .9206&\\
NLM$^{16}_{0.65}$ & 25.49 & .7580 & .8624 & .1043 & .9243 & .9213&\\
NLM$^{16}_{0.8}$ & 26.38 & .7942 & .8813 & \bf{.0977} & \bf{.9248} & \bf{.9223}&\\
NLM$^{16}_{0.9}$ & 27.11 & \bf{.8153} & \bf{.8910} & .1037 & .9168 & .9148&\\
NLM$^{16}_{1.0}$ & 27.42 & .8147 & .8870 & .1214 & .8971 & .8954&\\
\hline
\end{tabular}
}
\end{center}
\caption{Average image quality index for denoising the Kodak image dataset, corrupted by zero-mean Gaussian noise, different standard deviation $\sigma$ and different values of the offset parameter $o$. Bold numbers indicate the best results.}
\label{table:NLM_sigmaall}
\vspace{-1em}
\end{table}